\documentclass{article}

\usepackage[preprint, nonatbib]{neurips_2024}

\usepackage{footmisc}
\usepackage[utf8]{inputenc} 
\usepackage[T1]{fontenc}    
\usepackage{hyperref}       
\usepackage{url}            
\usepackage{booktabs}       
\usepackage{amsfonts}       
\usepackage{nicefrac}       
\usepackage{microtype}      
\usepackage{xcolor}         

\usepackage{graphicx}
\usepackage{wrapfig}
\usepackage{subfigure}

\usepackage{amsmath,amssymb}
\usepackage{float}
\usepackage{xspace}
\makeatletter
\DeclareRobustCommand\onedot{\futurelet\@let@token\@onedot}
\def\@onedot{\ifx\@let@token.\else.\null\fi\xspace}

\newcommand\figcaption{\def\@captype{figure}\caption}
\newcommand\tabcaption{\def\@captype{table}\caption}
\makeatother

\title{RepText: Rendering Visual Text via Replicating}

\author{
    Haofan Wang\footnotemark[2]\ , Yujia Xu, Yimeng Li, Junchen Li, Chaowei Zhang \\ \textbf{Jing Wang}, \textbf{Kejia Yang}, \textbf{Zhibo Chen} \\
    Shakker Labs, Liblib AI\\
    \footnotemark[2]\ \ Corresponding Author \\
    \texttt{haofanwang.ai@gmail.com} \\
    \textcolor{magenta}{\url{https://reptext.github.io}}
}

\begin{document}

\maketitle

\begin{abstract}

Although contemporary text-to-image generation models have achieved remarkable breakthroughs in producing visually appealing images, their capacity to generate precise and flexible typographic elements, especially non-Latin alphabets, remains constrained. This inherent limitation mainly stems from the fact that text encoders cannot effectively handle multilingual inputs or the biased distribution of multilingual data in the training set. In order to enable text rendering for specific language demands, some works adopt a dedicated text encoder or multilingual large language models to replace existing monolingual encoders, and retrain the model from scratch to enrich the base model with native rendering capabilities, but inevitably suffer from high resource consumption. The other works usually leverage auxiliary modules to encode text and glyphs while keeping the base model intact for controllable rendering. However, existing works are mostly built for UNet-based models instead of recent DiT-based models (SD3.5, FLUX), which limits their overall generation quality. To address these limitations, we start from an naive assumption that text understanding is only a sufficient condition for text rendering, but not a necessary condition. Based on this, we present \textbf{RepText}, which aims to empower pre-trained monolingual text-to-image generation models with the ability to accurately render, or more precisely, replicate, multilingual visual text in user-specified fonts, without the need to really understand them. Specifically, we adopt the setting from ControlNet and additionally integrate language agnostic glyph and position of rendered text to enable generating harmonized visual text, allowing users to customize text content, font and position on their needs. To improve accuracy, a text perceptual loss is employed along with the diffusion loss. Furthermore, to stabilize rendering process, at the inference phase, we directly initialize with noisy glyph latent instead of random initialization, and adopt region masks to restrict the feature injection to only the text region to avoid distortion of the background. We conducted extensive experiments to verify the effectiveness of our RepText relative to existing works, our approach outperforms existing open-source methods and achieves comparable results to native multi-language closed-source models. To be more fair, we also exhaustively discuss its limitations in the end. Code is available at \textcolor{magenta}{\url{https://github.com/Shakker-Labs/RepText}}.

\end{abstract}

\section{Introduction}
\label{sec:intro}

Visual text rendering aims to render text content visually instead of semantically that matches the text description in the generated image. It has a wide range of application scenarios, spanning graphic design (e.g., greeting cards, product poster) to natural scenes (e.g., car plate, shop sign, billboard). Although text-to-image generation models have demonstrated remarkable progress in generating visually
appealing and semantic aligned images, their ability to render precise textual elements remains suboptimal due to the inherent complexity of typographic elements (e.g., multilingual glyphs, font styles, spatial layouts). Previous works have demonstrated that using more powerful text encoders can directly enhance the model's text understanding and rendering capabilities. For example, the open source Stable Diffusion 3.5~\cite{esser2024sd3} and FLUX-dev~\cite{flux2024} use the T5~\cite{raffel2020exploring} encoder. Going further, using multilingual encoders or LLMs can enable the model to render multilingual text, such as the closed-source Seedream 3.0~\cite{jimeng}, Kolors 2.0~\cite{kolors}, and GPT4o~\cite{4o}. It is undeniable that improving the model’s ability to understand text plays an important role in rendering text. However, such models often need to be trained from scratch, which requires huge costs. In addition, they lack controllability and do not support users to precisely specify the spatial location of rendered text. 

To address these limitations, built on the top of text-to-image models, previous methods have used auxiliary control modules to restrict glyphs, such as GlyphControl~\cite{yang2023glyphcontrol}, AnyText~\cite{tuo2023anytext}, AnyText2~\cite{tuo2024anytext2}, GlyphDraw2~\cite{ma2025glyphdraw2}, ControlText~\cite{jiang2025controltext}, and JoyType~\cite{li2024joytype}. Other methods, such as TextDiffuser-2~\cite{chen2024textdiffuser} and Glyph-ByT5-v2~\cite{liu2024glyph}, implement text rendering by introducing new special tokens or introducing multi-language encoders. However, these solutions either cannot support multilingual text rendering or compromise generation quality when using outdated base models (SD1.5~\cite{rombach2022sd} or SDXL~\cite{podell2023sdxl}).

\begin{figure*}[htbp]
    \centering
    \includegraphics[trim=2cm 2cm 2cm 2cm, clip,width=1.0\linewidth]{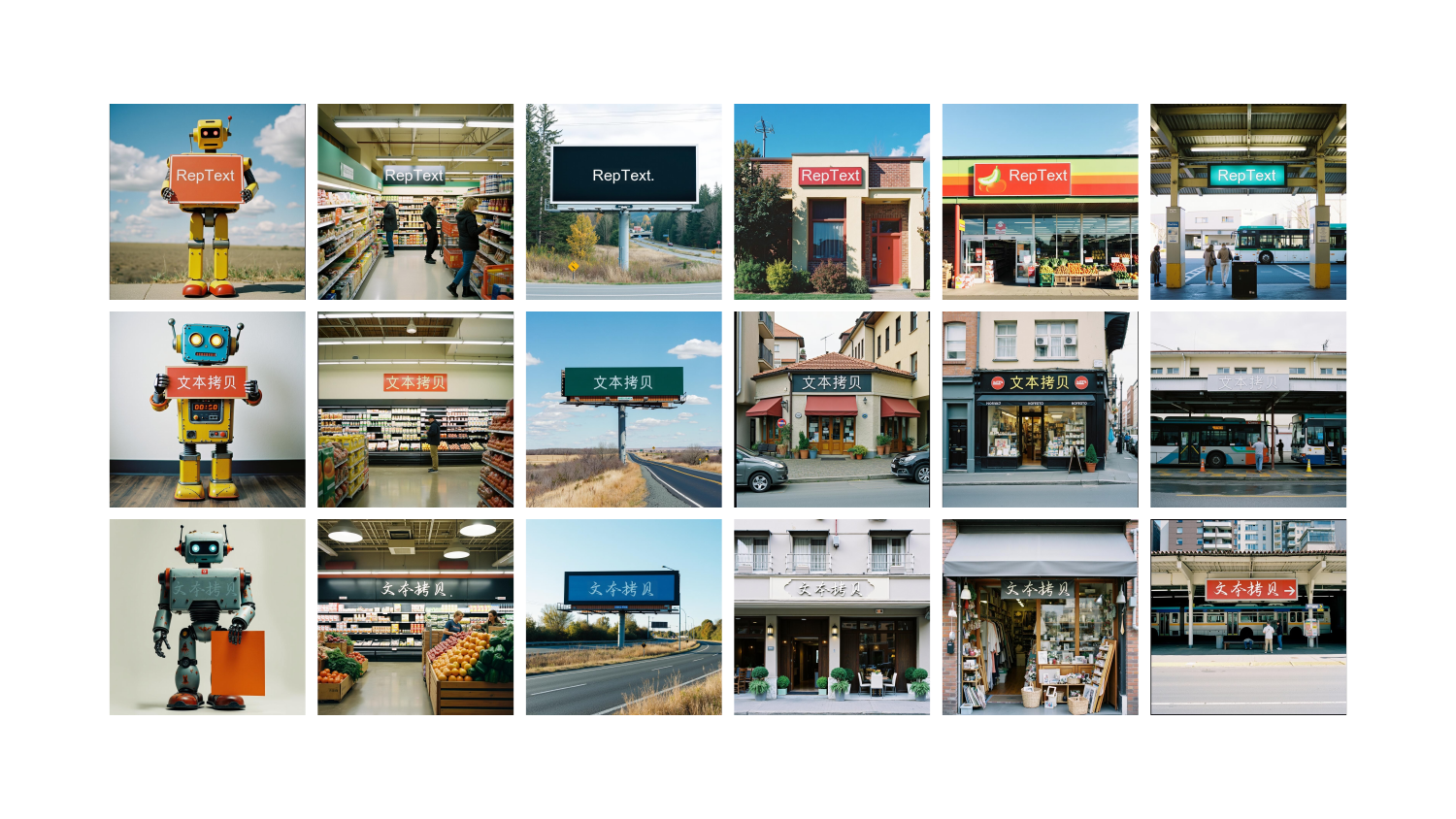}
    \caption{Illustrating of RepText generated samples for different text, languages and font conditions.}
    \label{fig:example}
\end{figure*}

In this paper, inspired by calligraphy copybooks, we start from a simple assumption that text understanding is only a sufficient condition for text rendering, but not a necessary condition. For example, when human children learn to write, they do not need to understand the specific meaning of each word, but only need to ensure the typographic accuracy. Based on this assumption, we introduce \textbf{RepText}, which aims to achieve text rendering based on the latest monolingual base models by replicating glyphs. Specifically, instead of using additional image or text encoders to understand words, we teach the model to replicate glyph via employ a text ControlNet~\cite{zhang2023adding} with canny and position images as conditions. Additionally, we innovatively introduce glyph latent replication in initialization to enhance text accuracy and support color control. Finally, a region masking scheme is adopted to ensure good generation quality and prevent the background area from being disturbed.

In summary, our contributions are threefold:

• We present RepText, an effective framework for controllable multilingual visual text rendering.

• We innovatively introduce glyph latent replication to improve typographic accuracy and enable color control. Besides, a region mask is adopted for good visual fidelity without background interference.

• Qualitative experiments show that our approach outperforms existing open-source methods and achieves comparable results to native multi-language closed-source models.

\section{Related Work}
\label{sec:related}

\subsection{Text-to-image Models.}
Over the past two years or so, text-to-image generation, especially diffusion-based generation has made incredible progress. From the open source Stable Diffusion 1.4~\cite{rombach2022sd} (2022.8), Stable Diffusion 1.5~\cite{rombach2022sd} (2022.10), Stable Diffusion XL~\cite{podell2023sdxl} (2023.7), Playground 2.5~\cite{li2024playground} (2024.2), Stable Diffusion 3.0~\cite{esser2024sd3} (2024.2), FLUX-dev~\cite{flux2024} (2024.8), Stable Diffusion 3.5~\cite{esser2024sd3} (2024.10), HiDream-L1~\cite{hidream} (2025.4), and the closed source Playground 3.0~\cite{liu2024playground} (2024.9), Recraft V3~\cite{recraft} (2024.10), Ideogram 3.0~\cite{ideogram} (2025.3), Reve Image~\cite{reve} (2025.3), GPT-4o~\cite{4o} (2025.3), Seedream 3.0~\cite{jimeng} (2025.4), Midjourney V7~\cite{mjv7} (2025.4), although not the latest released model is the best, there is no doubt that the iteration of image generation models is accelerating. In terms of visual model structure, from the early UNet to the current mainstream MMDiT~\cite{esser2024sd3}, the emergence of GPT-4o has brought hope to the paradigm of autoregression combined with diffusion models. In order to enhance text understanding, text encoders have gradually migrated from CLIP~\cite{radford2021learning} to T5~\cite{raffel2020exploring}, and there is a trend of further replacement with LLM or MLLM. These upgrades allow the model to generate images with better prompt following, higher aesthetic quality, higher resolution, and even have native multi-language understanding and rendering capabilities, as well as the ability to reference multiple concepts.

\subsection{Controllable Text-to-Image Generation.}
Controllable image generation usually refers to personalized generation or customized generation, that is, adding control signals for specified purposes and allowing text-to-image generation models to generate according to coarse-grained or fine-grained constraints, including spatial control, color, style, face, subject, etc. For example, ControlNet~\cite{zhang2023adding} and T2I-Adapter~\cite{mou2024t2i} are the most representative works for precise spatial control, which are conditioned on canny, depth maps, posture maps, etc. For subject-driven generation, classic works include training-based LoRA~\cite{hu2022lora}, Textual Inversion~\cite{gal2022image}, Dreambooth~\cite{ruiz2023dreambooth}, and recent works like IC-LoRA~\cite{huang2024context}, OminiControl~\cite{tan2024ominicontrol}, ACE++~\cite{mao2025ace++}, EasyControl~\cite{zhang2025easycontrol}, UNO~\cite{wu2025less}, and tuning-free IP-Adapter~\cite{ye2023ip}, InstantID~\cite{wang2024instantid}, InstantStyle~\cite{wang2024instantstyle, wang2024instantstyle-plus, xing2024csgo}, InstantCharacter~\cite{tao2025instantcharacter}, which focus on learning representations of specific style, identity, and subject from one or several images. In addition to plugin-controlled models, more and more one-for-all models, such as Show-O~\cite{xie2024show}, OmniGen~\cite{xiao2024omnigen}, Janus~\cite{wu2024janus}, and VisualCloze~\cite{li2025visualcloze}, have been proposed recently for unified image generation. There are also some works~\cite{chen2024training, chen2024region} supporting regional prompts.

\subsection{Visual Text Rendering.}
It is a consensus that text encoders play a vital role in the text understanding and rendering process. The current mainstream text rendering methods can be divided into two categories according to whether the text encoder can understand the text content to be rendered. Most of the current open source methods are based on Stable Diffusion 1.5~\cite{rombach2022sd} or Stable Diffusion XL~\cite{podell2023sdxl}, which means that although the CLIP tokenizer~\cite{radford2021learning} can encode English, it cannot handle multilingual input, and the accuracy of the rendered text is very poor. Therefore, in order to support accurate multilingual rendering, several works such as GlyphControl~\cite{yang2023glyphcontrol}, AnyText~\cite{tuo2023anytext}, AnyText2~\cite{tuo2024anytext2}, GlyphDraw2~\cite{ma2025glyphdraw2}, ControlText~\cite{jiang2025controltext}, and JoyType~\cite{li2024joytype} use a text ControlNet~\cite{zhang2023adding} to control the glyphs. Other works like TextDiffuser-2~\cite{chen2024textdiffuser} and Glyph-ByT5-v2~\cite{liu2024glyph} modify the design of the text encoder, such as introducing special tokens to refer to the text to be rendered, or using a tokenizer-free encoder such as ByT5~\cite{xue2022byt5}, but the generation quality of these works is often poor. In contrast, directly using a more powerful text encoder or multi-language encoder can bring more obvious improvements. For example, Stable Diffusion 3.5~\cite{esser2024sd3} and FLUX~\cite{flux2024} already have good English text rendering capabilities. We found that in this case, only a simple design is needed to achieve better controllable text rendering. Recent closed-source models, such as Kolors 2.0~\cite{kolors}, Seedream 3.0~\cite{jimeng}, and GPT4o~\cite{4o}, can render text more accurately and flexibly because the models have native multi-language understanding capabilities. However, due to the lack of open source and good multilingual base models, and the high cost of replacing text encoders and retraining, in this paper, our goal is to achieve text rendering based on the latest monolingual base models. Instead of modifying the text encoder side by making the model understand the text content to be rendered, we take another way to achieve text rendering by making the model reasonably replicate the text.

\section{Methods}
\label{sec:methods}

\subsection{Motivations}
We start from a simple philosophy, that is, whether understanding text is a necessary and sufficient condition for rendering text, especially text with simple strokes. We provide several toy examples. First, think back to how human children learn to write. Most children begin writing by scribbling and drawing, not really knowing what they are writing, but simply imitating what is already around them, then they begin to recognize words, and literacy and writing skills go hand in hand. A follow-up example is copybook, that contains examples of handwriting and blank space for learners to imitate. For some complex artistic fonts, especially non-Latin scripts such as Chinese calligraphy, imitating the glyphs may occur even earlier than recognizing the text. In short, although recognizing and understanding text is undoubtedly helpful for writing, we argue that writing can also start with imitation, or replicating, which should also be valid in rendering visual text in generative models.

Based on this naive assumption, we use the pre-trained ControlNet-Union~\cite{flux-cn-union-pro-2} that trained with canny edges on natural images to render text as preliminaries. As shown in Appendix Fig~\ref{fig:motivation}, it can already show some degree of replication, although there are obvious problems with the accuracy of the text, and the resulting degradation of the image quality. This motivate us to develop a method on the top of it that can replicate multilingual, multi-font text using existing monolingual text encoders.

\subsection{RepText}
\begin{figure*}[htbp]
    \centering
    \includegraphics[width=1.0\linewidth]{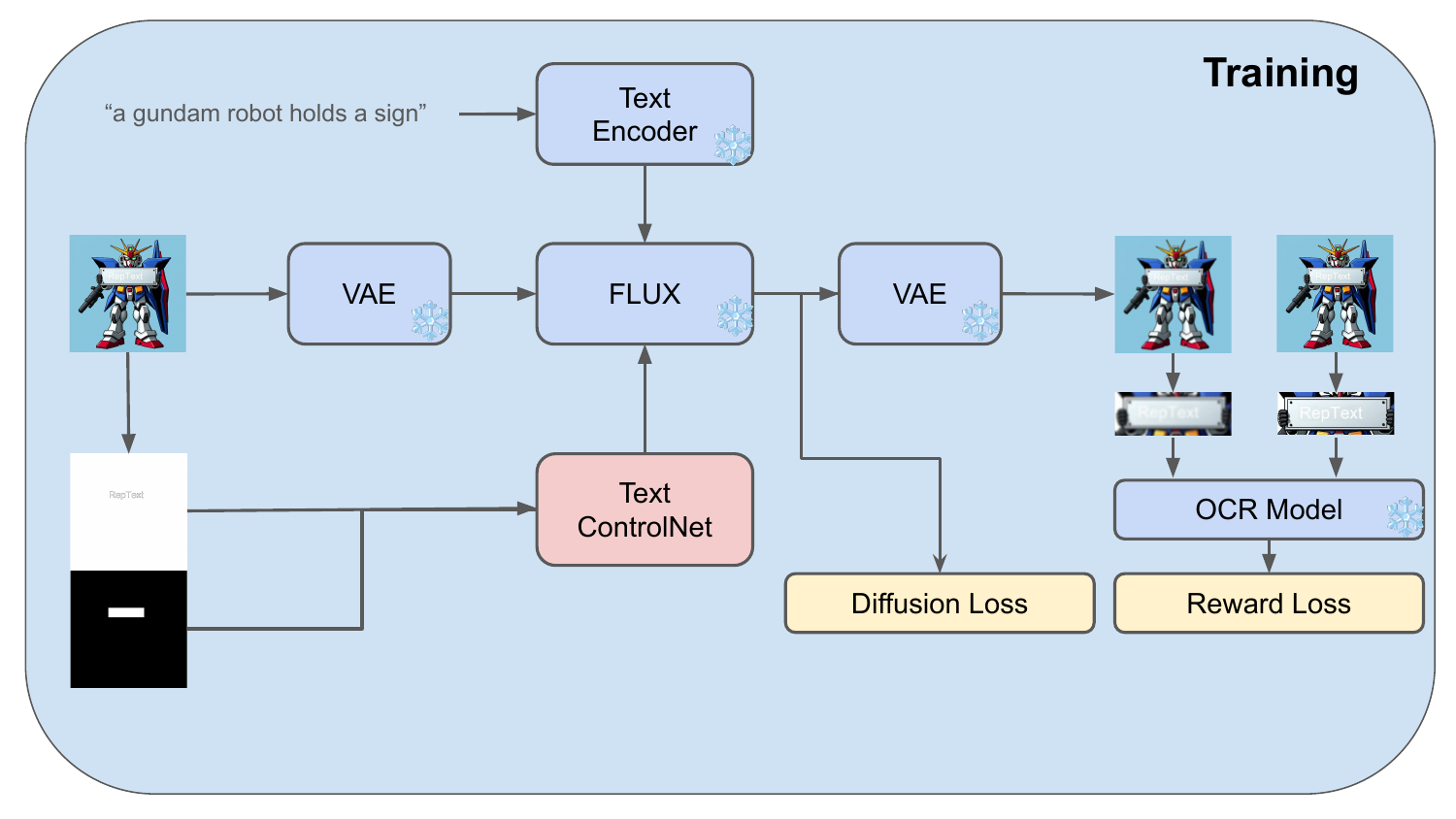}
    \caption{The training pipeline of RepText, where we use both fine-grained canny edge and position mask as conditions to train text ControlNet, and further adopt a text perceptual loss.}
    \label{fig:train}
\end{figure*}

\textbf{Framework.} As shown in Fig \ref{fig:train}, the RepText is a ControlNet-like framework and mostly inspired by GlyphControl~\cite{yang2023glyphcontrol} and JoyTypes~\cite{li2024joytype}. To incorporate fine-grained glyph information and enable multilingual rendering, instead of directly using rendered glyph images as GlyphControl~\cite{yang2023glyphcontrol} where they depend on text encoder to understand the semantic meaning of the words, we use stronger text hint, canny edge extracting from images, additionally, to provide location information, we also use auxiliary position image to assist text rendering. The canny and position images are processed by the VAE encoder separately and concatenated over channel dimension before feeding into ControlNet branch. The text contents to be rendered are not manually added into prompts.

To improve the accuracy of text generation, we further adopt a text perceptual loss as AnyText~\cite{tuo2023anytext}. Specifically, in the training phase, given the predicted noise $\epsilon_{t}$, current timestep $t$ and noisy latent image $z_{t}$, we can directly predict $z_{0}$ as described~\cite{ho2020denoising}. Then, we use the VAE decoder to obtain the approximated $x_0^{'}$ in pixel space. As we already have ground-truth annotations of text lines, we can accurately localize text regions from $x_0^{'}$ and $x_0$, and use the cropped text images as inputs for the OCR model. Following AnyText~\cite{tuo2023anytext}, we also employ the PP-OCRv3 model~\cite{li2022pp}.

The text perceptual loss is expressed as

\begin{equation}
    L_{reward} = \sum_{p}\frac{1}{hw}\sum_{h,w}||m_p - m_p^{'}||^{2}_{2}
\end{equation}

where $m_p$, $m_p^{'}$ $\in \mathbb{R}^{h\times w \times c}$ are feature maps before the last fully connected layer of OCR model that represent the textual information in $x_0$ and $x_0^{'}$ at position $p$. This Mean Squared Error (MSE) loss is used to improve the recognizability of generated text.

The overall objective for training is formulated as

\begin{equation}
    L = L_{denoise} + \lambda \times L_{reward}
\end{equation}

where $\lambda$ is the scaling factor that adjusts the weights of reward OCR loss and denoising loss, and is empirically set to a small value such as 0.10 or 0.05.

\begin{figure*}[htbp]
    \centering
    \includegraphics[width=1.0\linewidth]{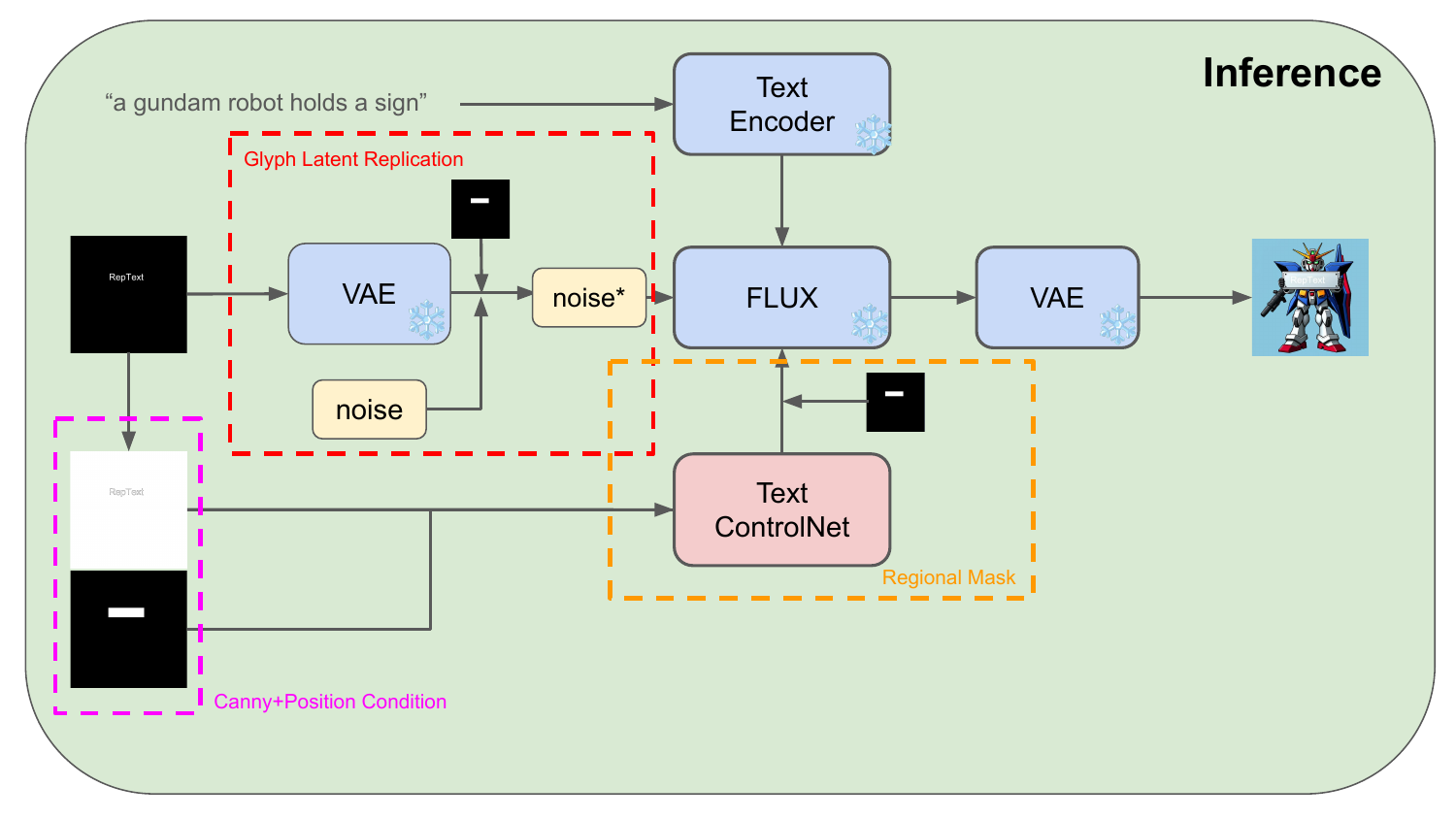}
    \caption{The inference framework of RepText with highlighted strategies: (1) Replicating from noise-free glyph latent, which improves text accuracy and enables color control. (2) Adopt regional mask for text regions, which avoids interference with non-text areas and ensures overall quality.}
    \label{fig:infer}
\end{figure*}

\textbf{Inference Strategy.} In the inference phase, we introduce several key techniques as shown in Fig ~\ref{fig:infer} to stabilize and improve text rendering performance.

\textit{Replicating from glyph latent.} Inspired by copybook, we initialize from noise-free glyph latent instead of random Gaussian noise, in other words replicate, to provide glyph guidance information at the beginning of the denoising steps. Only the text regions of noise-free glyph latent will be replicated and pasted back into random noise. We find that such a simple step plays an important role in improving the accuracy of rendered text. Benefiting from this design, RepText further allows users to specify text color without implicitly encoding color information through learnable layers. In our implementation, we find that directly copying and pasting would lead to a significant degradation of image quality because the noise-free area is not a Gaussian noise. Therefore, we introduce a weight coefficient to control the influence of glyph latent. The initialized latent $z_{T}$ is defined as follows.

\[
z_{T} = 
\begin{cases} 
\lambda_{1} \times \mathcal{N}(0, 1) + \lambda_{2} \times z_{0}^{*} & \text{if } (x,y) \in p \\
\mathcal{N}(0, 1) & \text{if } (x,y) \notin p
\end{cases}
\tag{3}
\]

where the $z_{0}^{*}$ is the noise-free glyph latent encoded by VAE, $\lambda_{1}$ and $\lambda_{2}$ are strength coefficients for random noise and glyph latent. Only text regions $p$ will be replicated. Please note that $z_{0}^{*}$ can also be obtained via inversion technique, but it shows small difference.

\textit{Regional mask for text regions.} Traditional ControlNet usually uses global hints as conditions, that is, both canny and depth are calculated on the entire image, while in our case, the conditional image is sparse and only the text area is valid. Therefore, in order to avoid interference with non-text areas during the denoising process, we additionally use regional masks to truncate the output of ControlNet, where the regional masks are binary and the text regions denoted by the bounding box are set to 1.

\begin{figure*}[htbp]
    \centering
    \includegraphics[trim=8cm 0cm 8cm 0cm, clip,width=1.0\textwidth]{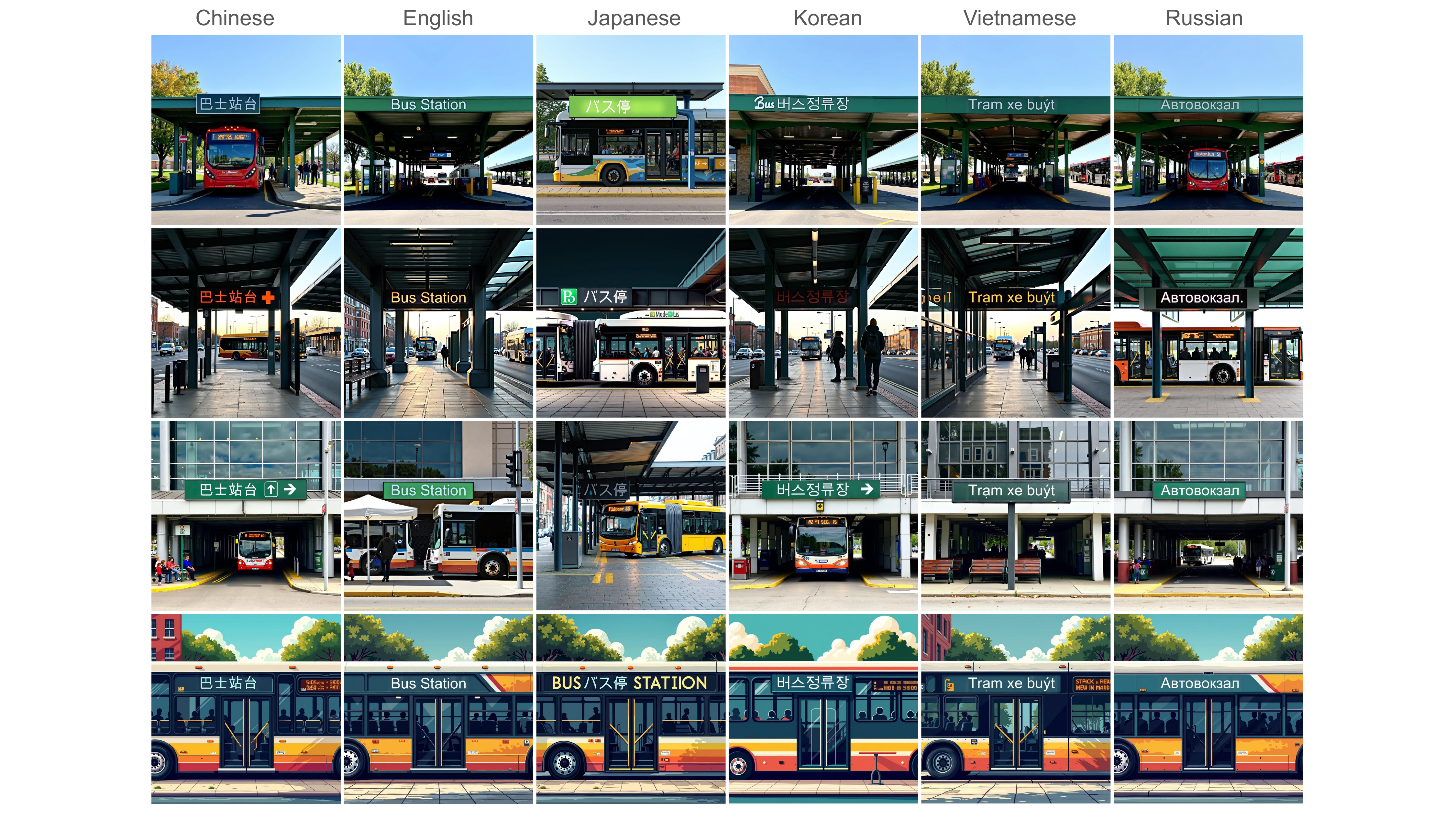}
    \caption{RepText can render multilingual texts by replicating glyph condition.}
    \label{fig:multilingual}
\end{figure*}

\section{Experiments}
\label{sec:exp}

\subsection{Implementation Details}
We implemented our method on top of a widely adopted open-sourced text-to-image model FLUX-dev~\cite{flux2024}, the text ControlNet branch consists of 6 double blocks and 0 single block following ControlNet-Union-Pro-2.0~\cite{flux-cn-union-pro-2} and is initialized from FLUX-dev. We use Anytext-3M~\cite{tuo2023anytext} as pre-training dataset (all images are in 512x512). The training resolution is set to 512, the AdamW optimizer is used with a learning rate of 2e-5 and a batch size of 256. The OCR loss scale is set to 0.05, and we set the text drop ratio to 0.3. Besides, we collect a high-quality image dataset containing 10K images for fine-tuning, all of which are natural images such as road signs, store signs, etc., rather than synthetic. When fine-tuning, we enable buckets to train on different aspect ratios, decrease learning rate to 5e-6, and increase OCR loss scale to 0.10 and text drop ratio to 0.4. In the training phase, we render position images based on annotations, the canny images are extracted from masked images where the non text regions are set to 0. In the inference phase, users are allowed to render their text on a blank image and use this glyph image to generate canny and position conditions. We set $\lambda_1$ and $\lambda_2$ to 0.9 and 0.1 empirically.

\begin{figure*}[htbp]
    \centering
    \includegraphics[trim=1cm 3cm 1cm 3cm, clip,width=1.0\linewidth]{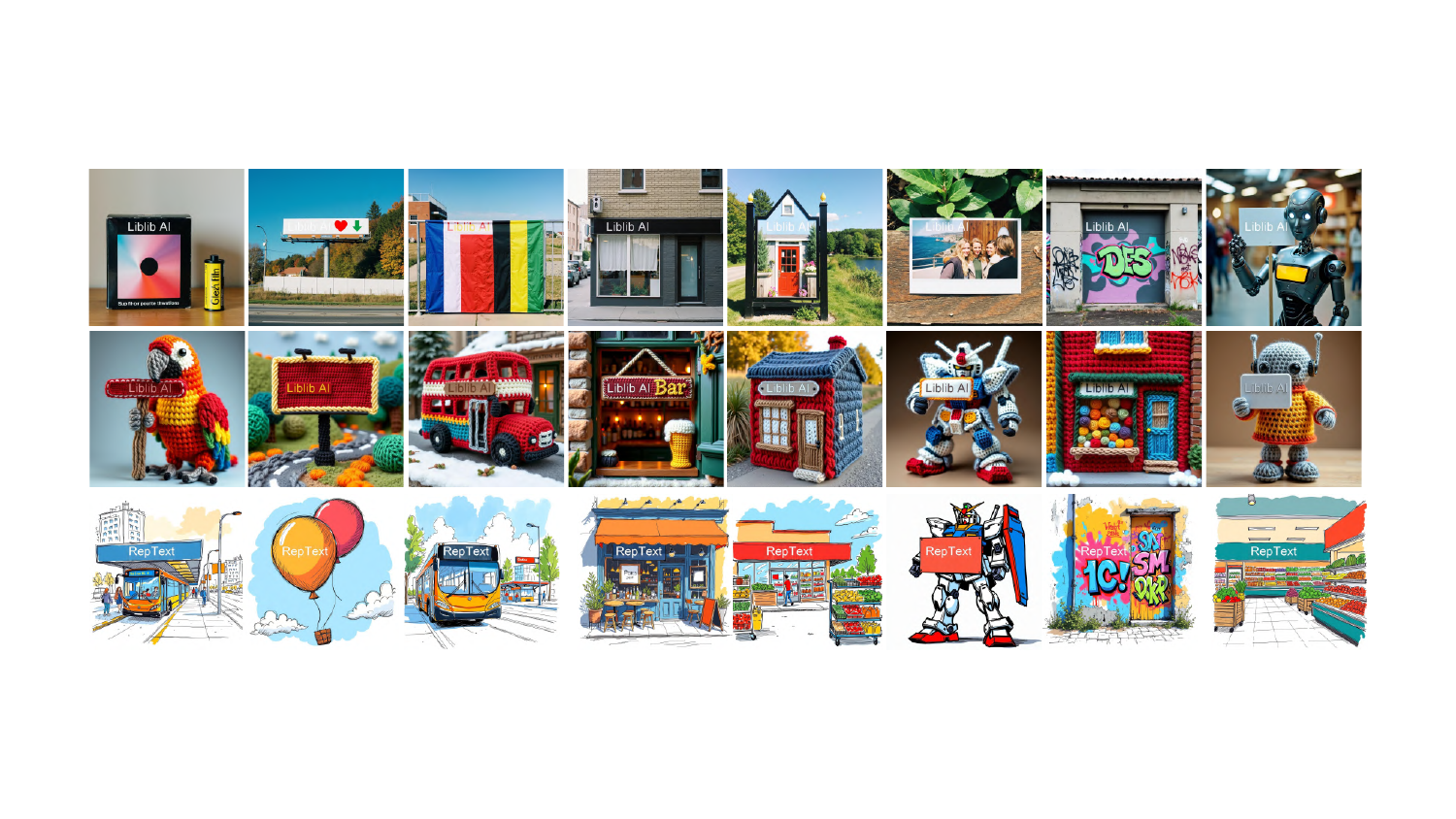}
    \caption{Illustrating of RepText's compatibility to community LoRAs. From top to bottom, they are FilmPortrait, wool yarn art and sketched style respectively.}
    \label{fig:lora}
\end{figure*}\textbf{}

\subsection{Qualitative Results}

Qualitative evaluations are performed for different scenes, including multiple languages especially non-Latin, multiple fonts, multiple colors, and multiple lines. The multilingual results are shown in Fig \ref{fig:multilingual}. The other results can be found in Appendix Fig \ref{fig:multi-fonts}, Fig \ref{fig:multi-colors}, and Fig \ref{fig:multi-lines} respectively to save pages. Benifit from our glyph replication, RepText can render accurate and controllable text content. More generated samples can be found in Appedix Fig~\ref{fig:poster} and Fig~\ref{fig:more-examples}

\subsection{Comparison to Previous Methods}

\textbf{Baseline methods.} For a comprehensive comparison, we compare with existing methods with monolingual or multilingual text rendering ability, including available open source and closed source models. For open source models, we use their official codes for inference unless otherwise specified. For closed-source models, we use their products or APIs for inference.

\textit{Monolingual.} We compare with the open source Stable Diffusion 3.5 large~\cite{esser2024sd3}, FLUX-dev~\cite{flux2024} and HiDream-I1-Dev~\cite{hidream}, and the close source FLUX 1.1 Pro Ultra~\cite{flux-1.1-pro-ultra}, Ideogram 3.0~\cite{ideogram}, Reve Image (Halfmoon)~\cite{reve} and Recraft V3~\cite{recraft}. Besides, we also compare with available open source methods that focus on controllable visual text rendering, including TextDiffuser~\cite{chen2023textdiffuser}, TextDiffuser2~\cite{chen2024textdiffuser}, and GlyphControl~\cite{yang2023glyphcontrol}. Specially, we re-implement GlyphControl on FLUX-dev. For Recraft V3~\cite{recraft}, we use its online Frame function which is based on TextDiffuser2~\cite{chen2024textdiffuser} for controllable rendering. The result is in Appendix Fig~\ref{fig:compare-monolingual}. For rendering Latin text, since the base model itself has excellent understanding and rendering capabilities, RepText acts more like a position guide and allows users to specify fonts. Explicitly adding the text content (English) to be rendered in the prompt can also help but is not used in our experiments.

\textit{Multilingual.} We compare with the open source Kolors 1.0~\cite{kolors} and Cogview4~\cite{cogview4}, and the close source Kolors 1.5~\cite{kolors}, Gemini Flash 2.0~\cite{gemini}, Wan2.1 Pro~\cite{wan}, GPT-4o~\cite{4o}, Seedream 3.0~\cite{jimeng} and Kolors 2.0~\cite{kolors}. Please note that although Hunyuan-DiT~\cite{li2024hunyuandit} uses mT5~\cite{xue2020mt5} as text encoder, it doesn't support multilingual text rendering. The result is in Appendix Fig ~\ref{fig:compare-multilingual}. Compared with open source methods, we have significant advantages in text accuracy and image quality. Compared with closed-source models using multilingual text encoders, we have better controllability. However, it must be admitted that due to their native multilingual understanding capabilities, the current most advanced text-to-image models like GPT-4o, Seedream 3.0, and Kolors 2.0 are more flexible in rendering text content than we are.

\subsection{Compatibility to Existing Works}

To demonstrate the compatibility and effectiveness of our approach, we equip RepText with common adopted plugin models including style LoRAs, other ControlNets and IP-Adapter.

\textbf{LoRAs}. We use three open source LoRAs available on HuggingFace. Specifically, we select FilmPortrait\footnote{https://huggingface.co/Shakker-Labs/FilmPortrait} which provides film texture, FLUX.1-dev-LoRA-MiaoKa-Yarn-World\footnote{https://huggingface.co/Shakker-Labs/FLUX.1-dev-LoRA-MiaoKa-Yarn-World} which creates wool yarn art, and FLUX.1-dev-LoRA-Children-Simple-Sketch\footnote{https://huggingface.co/Shakker-Labs/FLUX.1-dev-LoRA-Children-Simple-Sketch} for sketched style. As shown in Fig \ref{fig:lora}, our work is fully compatible to community LoRAs for stylization.

\textbf{Other ControlNets}. We use ControlNet-Union-Pro-2.0~\cite{flux-cn-union-pro-2} and ControlNet-Inpainting~\cite{flux-cn-inpaint} to achieve spatial control and text editing. The results are shown in Appendix Fig \ref{fig:controlnet}.

\textbf{IP-Adapter}. We use FLUX.1-dev-IP-Adapter~\cite{flux-ipa} as example. As shown in Appendix Fig~\ref{fig:ipa}, our method can be used together with IP-Adapter.

\subsection{Ablation Studies}

\textbf{Choice of ControlNet condition.} We conducted experiments to analyze the impact of different ControlNet conditions. In the case of position only condition, it only provides position guidance, and in the case of Canny only condition, we can render the corresponding text, but the accuracy and harmony are limited, while with the joint Canny and position conditions, we can accurately render harmonious text. The result is in Appendix Fig~\ref{fig:ablation-choice-cn}.

\textbf{Effect of Glyph Latent Replication.} As shown in Appedix Fig~\ref{fig:ablation-glyph-latent} (left), initializing from glyph latent improves typographic accuracy at no cost. Additionally, it allows the user to specify colors without relying on an additional color encoder as shown in Appedix Fig~\ref{fig:ablation-glyph-latent} (right).

\textbf{Effect of Regional Mask.} Different from other ControlNets where the control signals are usually global and dense, text is a local and sparse control. We find that introducing region masks in inference phase helps improve the quality of non-text background as shown in Appendix Fig~\ref{fig:ablation-region-mask}.

\subsection{Limitations and Future Works}

\textbf{Typical Failure Cases}. Although RepText shows good text rendering capabilities and compatibility, it still has some limitations that stem from its own lack of understanding of text. We discuss several common bad cases as below.

\textit{Disharmony with the scene.} Although the training dataset contains a lot of text data from natural scenes, such as road signs, the text encoder (T5-XXL) itself does not understand the text content that needs to be rendered (even the text content is added into prompt), referring to non-Latin text, so sometimes the text is stiffly pasted on the generated image as a signature or watermark, resulting in disharmony with the scene as shown in Appendix Fig \ref{fig:bad-cases} (a).

\textit{Limited text accuracy.} For texts with complex strokes such as Tibetan or small fonts, even with our framework, the rendering accuracy is still poor as shown in Appendix Fig \ref{fig:bad-cases} (b). One of the reasons is that the control conditions is not precise enough, and the current compression ratio of VAE also leads to poor rendering of small characters.

\textit{Render extra texts.} We find that even with the regional mask, some extra text still appeare in the non-rendered text area as shown in Appendix Fig \ref{fig:bad-cases} (c), and these text are usually meaningless, unrecognizable or repeated.

\textit{Limited text diversity.} Limited by the text encoder, we have to utilize extra conditions and cannot flexibly control text properties through prompt, including its position, color, material, etc.

\textit{Not support precise color control.} While initializing from the glyph potential allows for coarse color control, it cannot strictly render fine-grained colors, limiting its application in real-world scenarios.

\textit{Lack of distortion and perspective.} Since the text content is completely controlled by front-view glyphs, limited by front-end rendering, it is inconvenient and difficult to flexibly generate text with deformation and perspective, and it is also difficult to generate some stylized text with distortion.

\textbf{Future Works}. As we have stated in previous sections, we acknowledge that the most flexible and effective way to render text is to let the model understand the specific meaning of each word, that is, to use a multilingual text encoder or MLLM, so as to achieve text rendering in natural scenes or poster scenes. The main concern is, besides replacing the text encoder and retraining it from scratch, is there a low-cost way, using fewer training parameters and training data, to enable existing text-to-image models to recognize and render text in different languages without compromising the original generation capabilities? For example, MetaQuery~\cite{pan2025transfer} reveals that MLLM's understanding and reasoning capabilities can be used to augment image generation when both the MLLM backbone and Diffusion backbone are kept frozen but only a lightweight connector is trained, such an approach may also be applicable to visual text rendering.

\section{Conclusion}
In this work, motivated by calligraphy copybooks, we present a simple yet effective framework \textbf{RepText} for controllable multilingual visual text rendering. We enable pre-trained monolingual text-to-image models with comparable capacity to generate legible visual text in different languages, fonts, and colors. Specifically, instead of using additional image or text encoders to understand words, we teach the model to replicate glyph via employ a text ControlNet with canny and position images as conditions. Additionally, we innovatively introduce glyph latent replication to enhance text accuracy and support color control. Finally, a region masking scheme is adopted to ensure good generation quality without interference from text information. Our approach outperforms existing open-source methods and achieves comparable results to native multi-language closed-source models. Next, we will explore how to efficiently enable monolingual models to understand multiple languages, thereby further improving the flexibility and accuracy of text rendering.

\bibliographystyle{splncs04}
\bibliography{example_paper}

\appendix

\clearpage
\newpage

\section{Supplementary Details}

\subsection{Preliminary Results}
\begin{figure*}[htbp]
    \centering
    \includegraphics[trim=0cm 2.5cm 0cm 2.5cm, clip,width=1.0\linewidth]{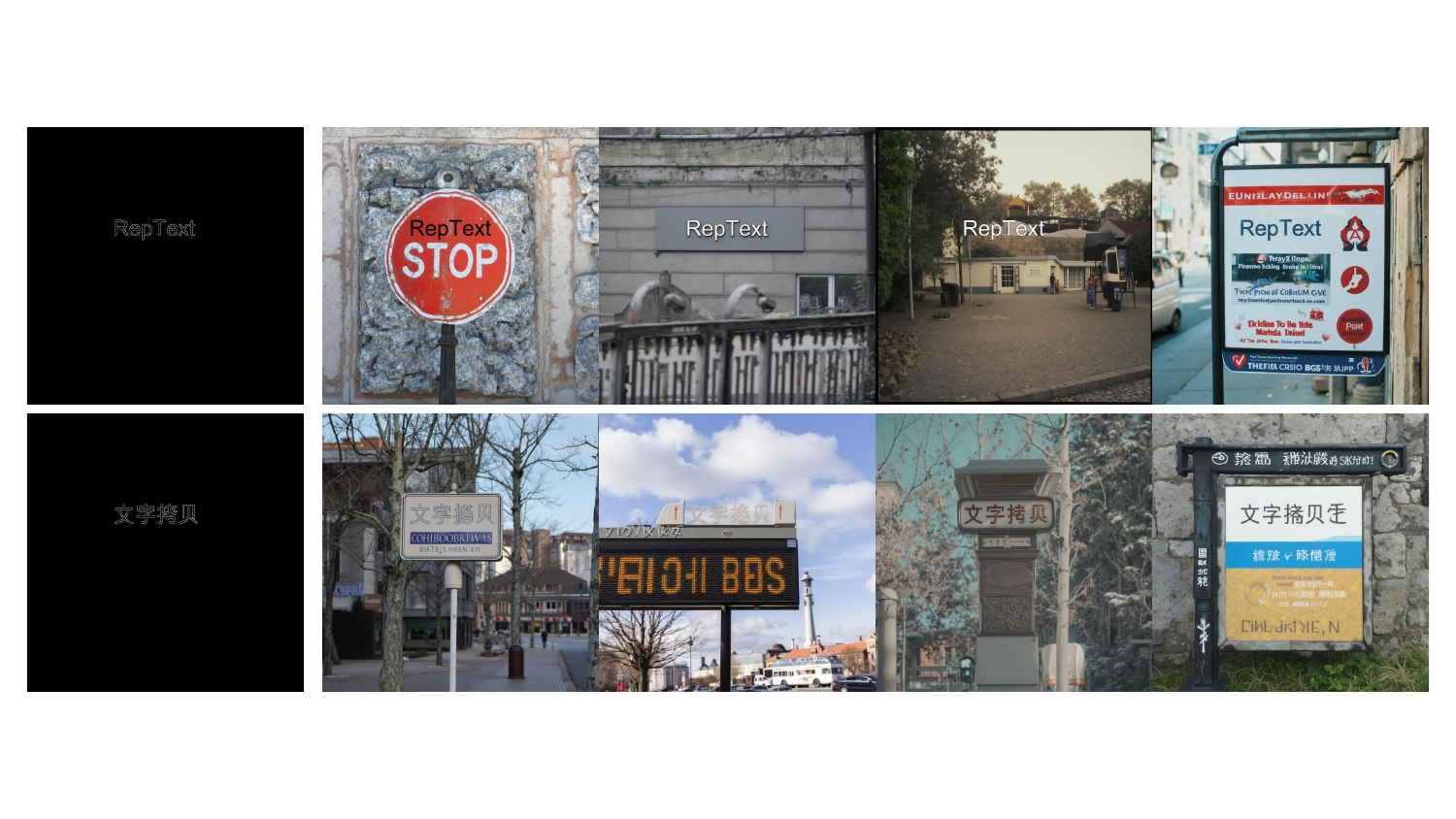}
    \caption{Although ControlNet (Canny) is not trained specifically on visual text datasets, it can still reasonably render text by replicating the given glyph edges to some extent.}
    \label{fig:motivation}
\end{figure*}

\subsection{Qualitative Results}

\begin{figure*}[htbp]
    \centering
    \includegraphics[trim=8cm 5cm 8cm 5cm, clip,width=1.0\linewidth]{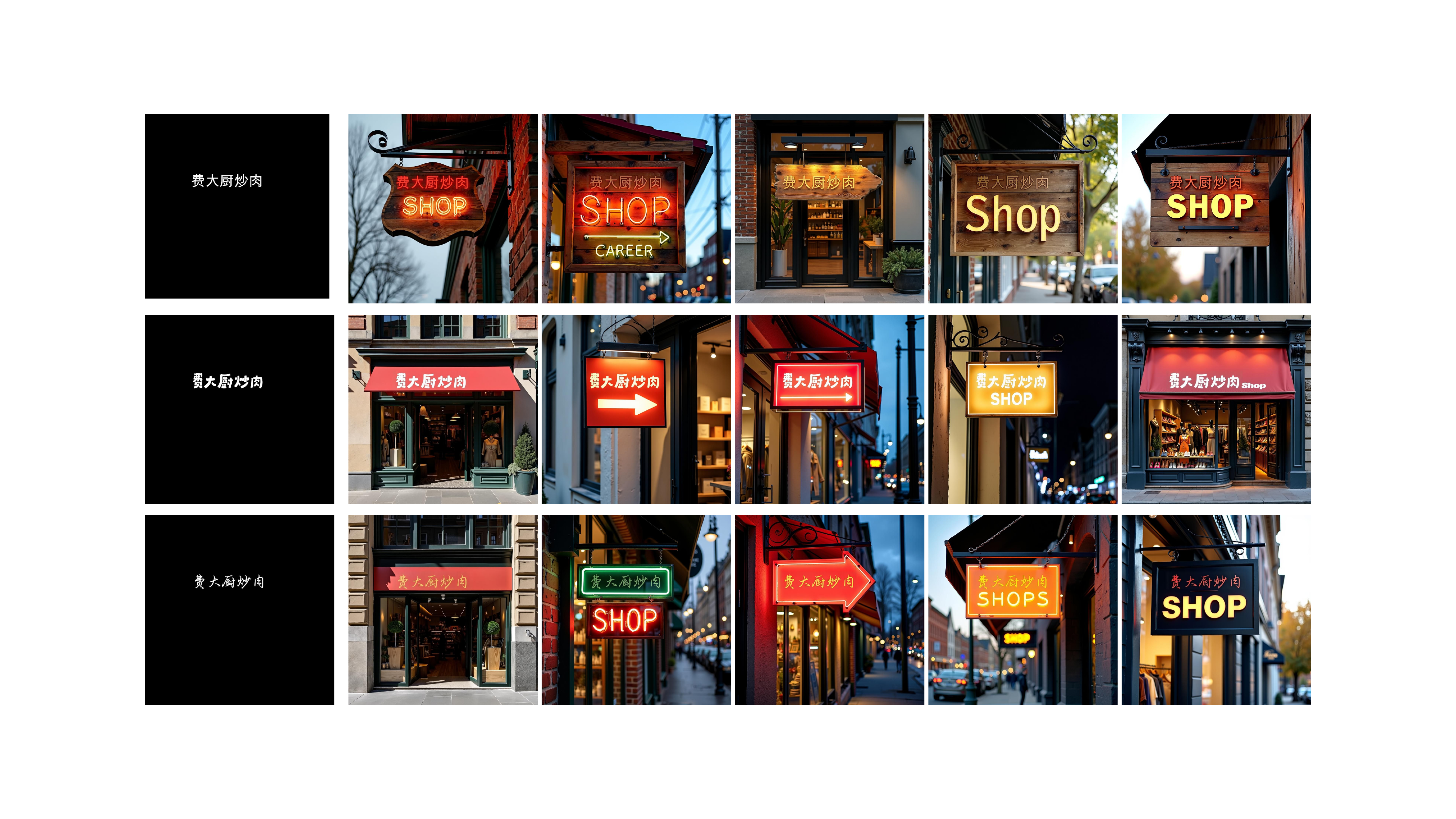}
    \caption{RepText can render texts with use-specified fonts by replicating glyph condition.}
    \label{fig:multi-fonts}
\end{figure*}

\begin{figure*}[htbp]
    \centering
    \includegraphics[trim=10cm 7cm 10cm 7cm, clip,width=1.0\linewidth]{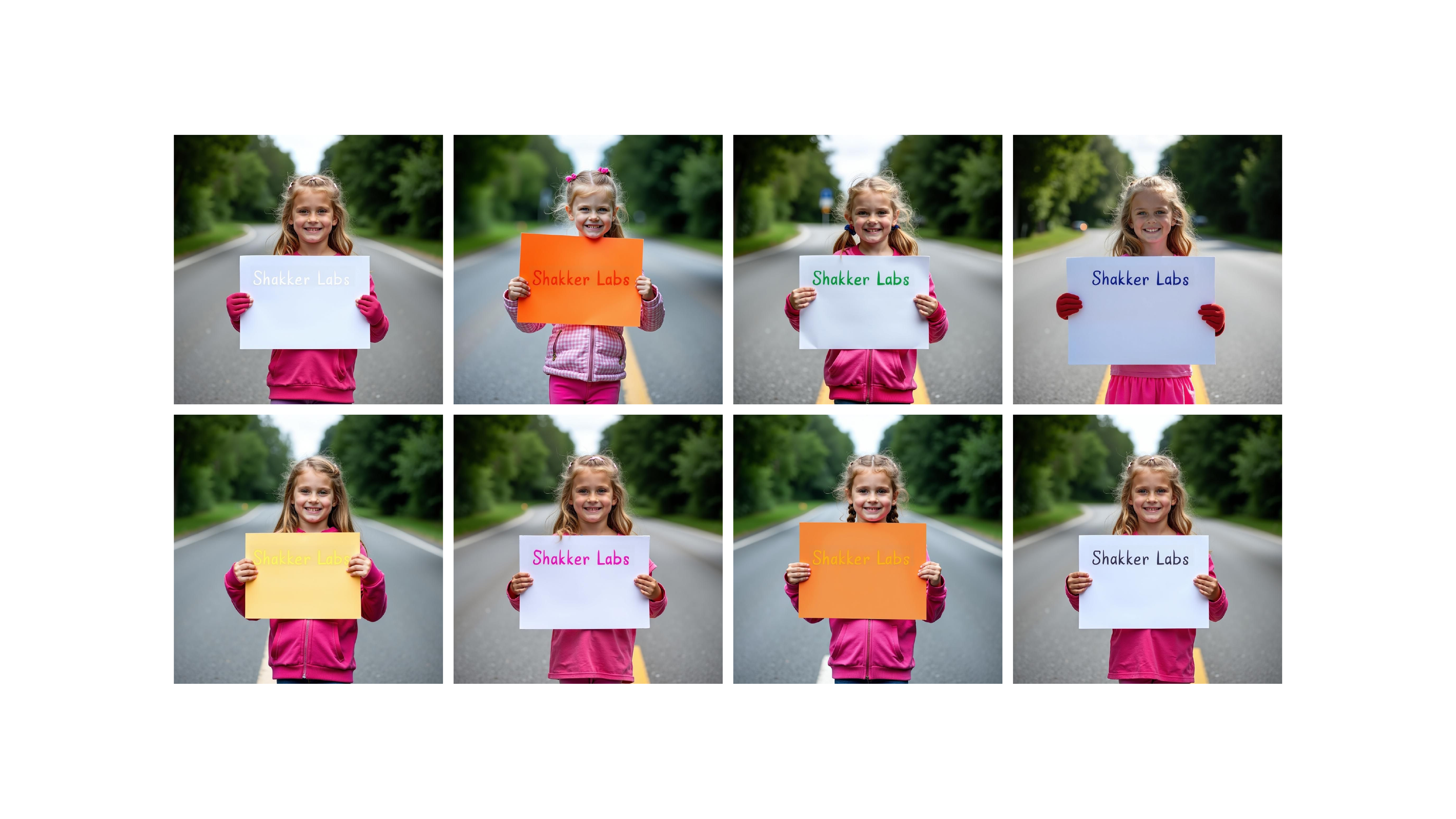}
    \caption{RepText can render texts with use-specified colors by initializing from glyph latent.}
    \label{fig:multi-colors}
\end{figure*}

\begin{figure*}[htbp]
    \centering
    \includegraphics[trim=12cm 0cm 12cm 0cm, clip,width=1.0\linewidth]{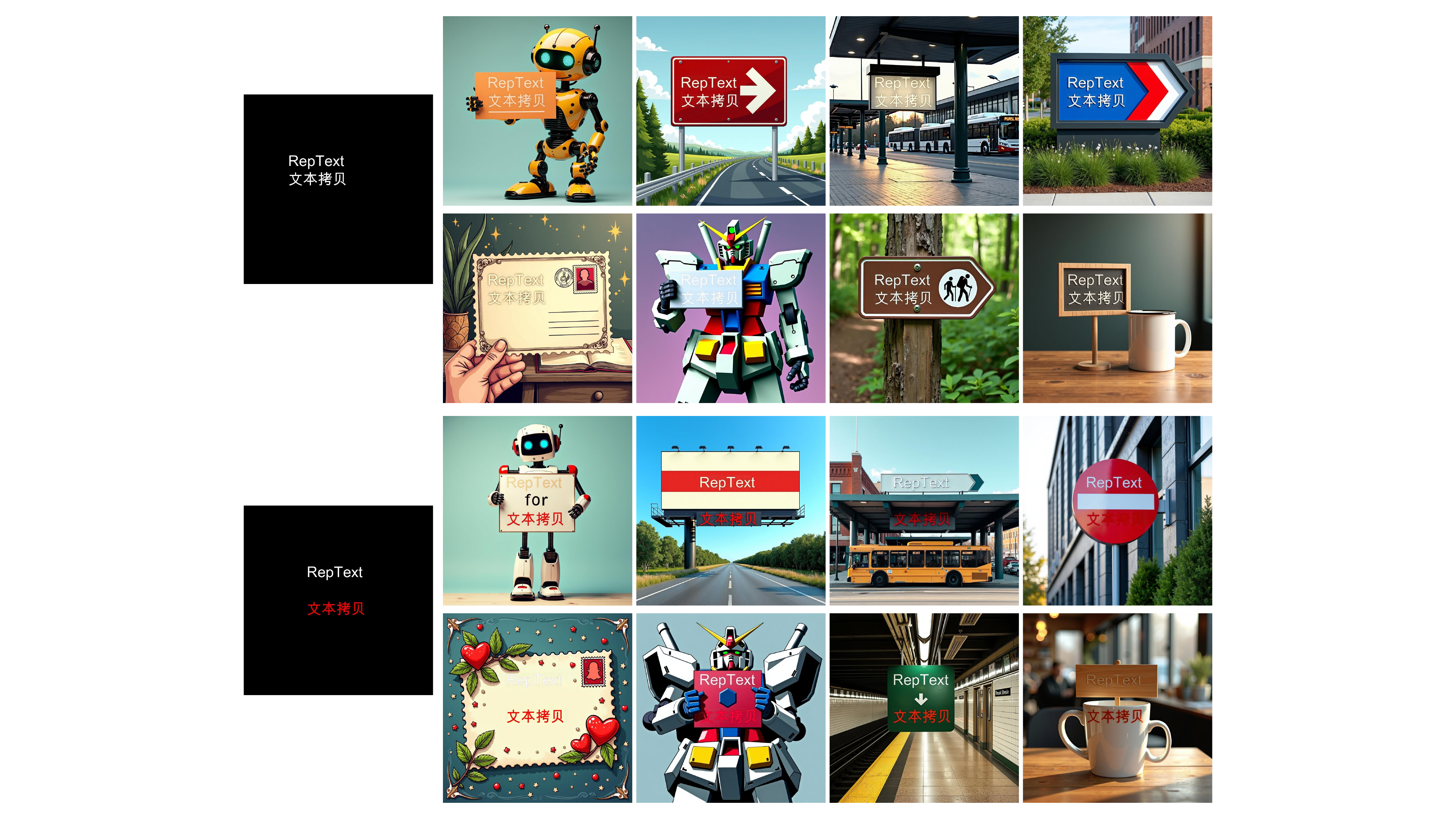}
    \caption{RepText can render multi-lines texts.}
    \label{fig:multi-lines}
\end{figure*}

\clearpage
\newpage
\subsection{More Qualitative Results}

\begin{figure*}[htbp]
    \centering
    \includegraphics[trim=1cm 1cm 1cm 1cm, clip,width=1\linewidth]{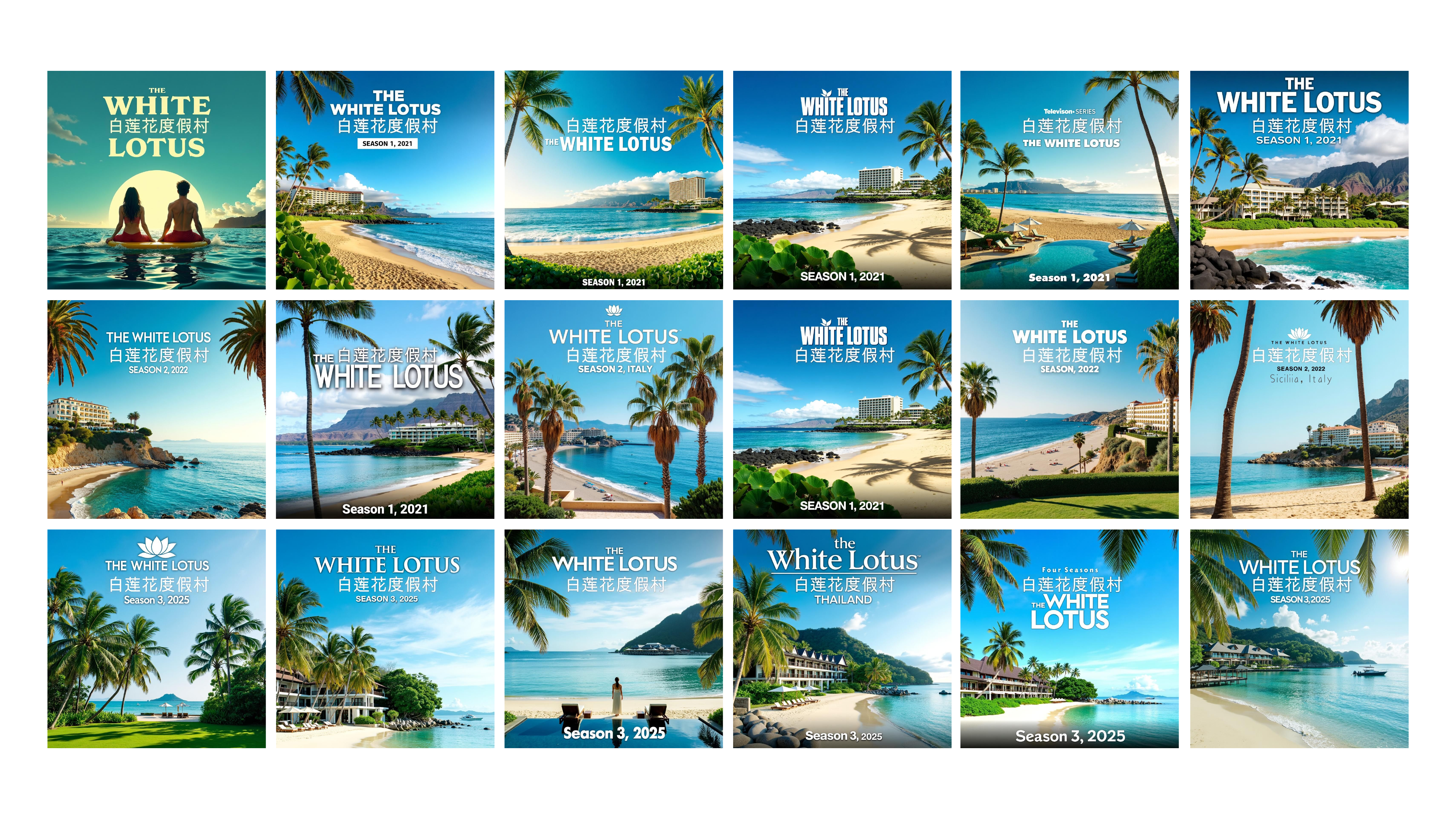}
    \includegraphics[trim=1cm 1cm 1cm 1cm, clip,width=1\linewidth]{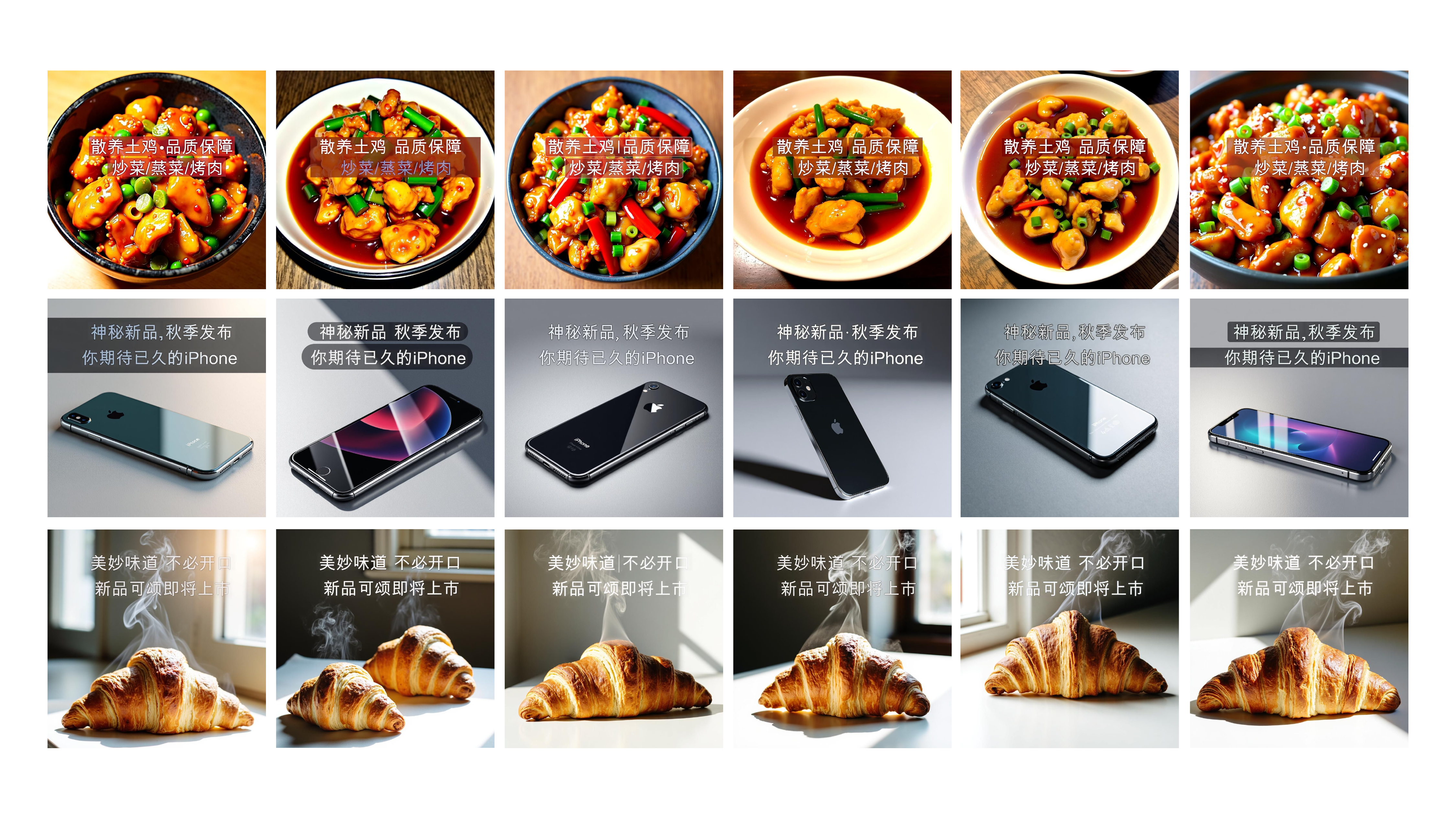}
    \caption{RepText can also synthesize movie or product posters using multilingual rendering capabilities and the generative model’s native Latin rendering capabilities.}
    \label{fig:poster}
\end{figure*}

\begin{figure*}[ht]
    \centering
    \includegraphics[trim=2cm 1cm 2cm 1cm, clip,width=1.0\linewidth]{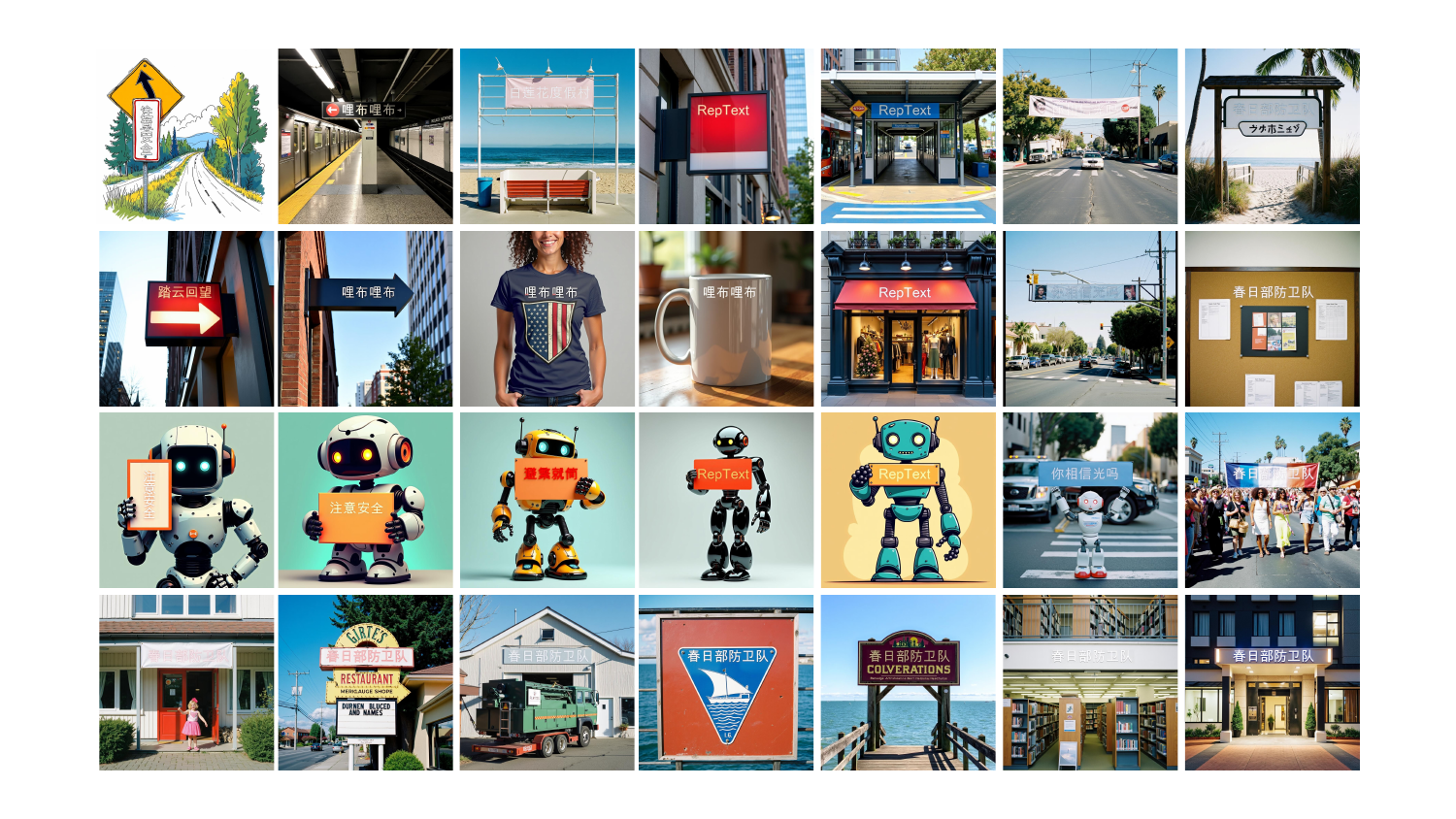}
    \includegraphics[trim=2cm 1cm 2cm 1cm, clip,width=1.0\linewidth]{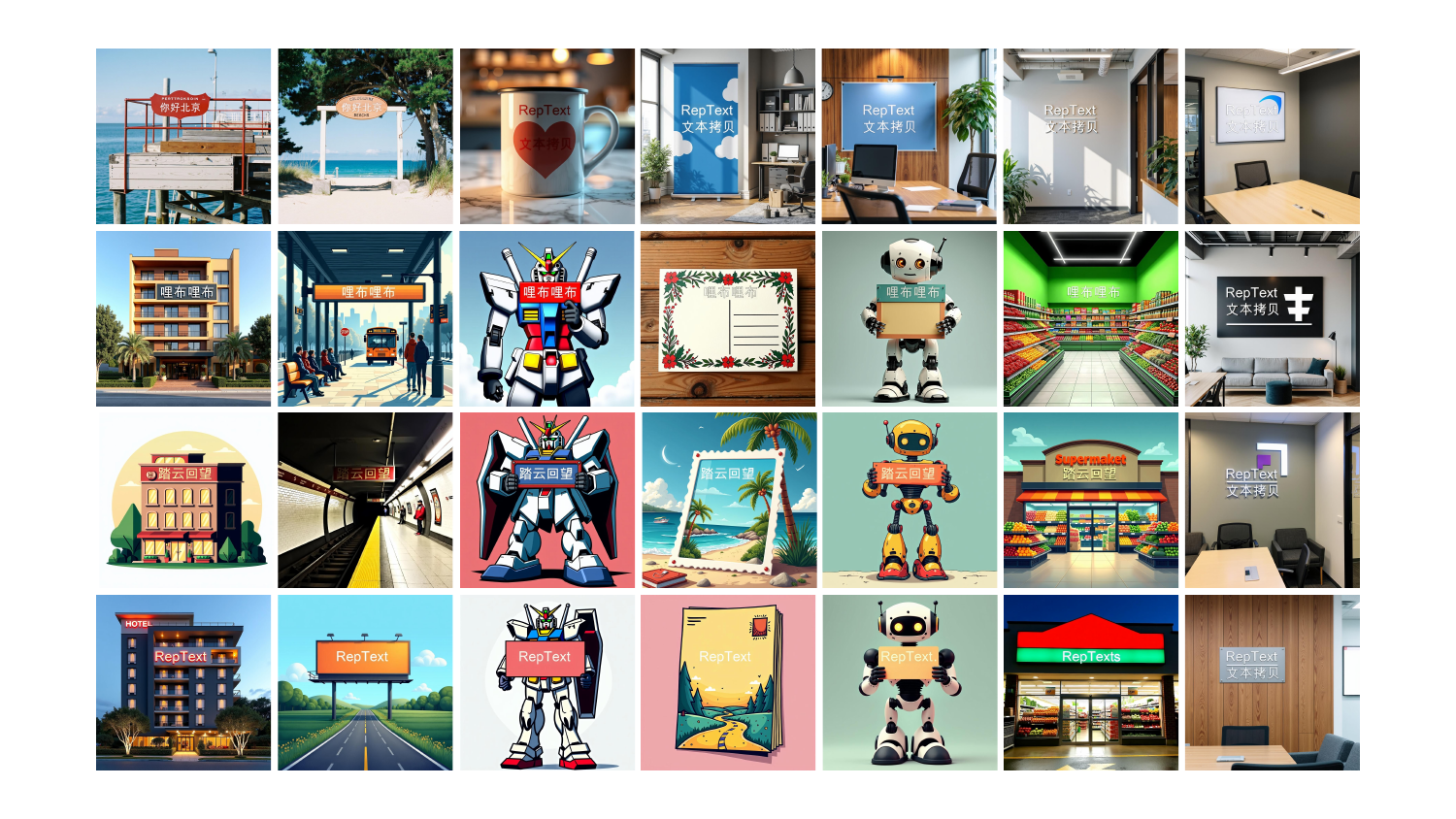}
    \caption{Supplementary samples generated by RepText.}
    \label{fig:more-examples}
\end{figure*}

Fig~\ref{fig:poster} shows our ability to generate TV and product posters, and Fig~\ref{fig:more-examples} shows more supplementary samples in natural scenes generated by RepText.

\newpage
\subsection{Comparison with Other Works}
\begin{figure*}[htbp]
    \centering
    \includegraphics[trim=8cm 0cm 8cm 0cm, clip,width=1.0\textwidth]{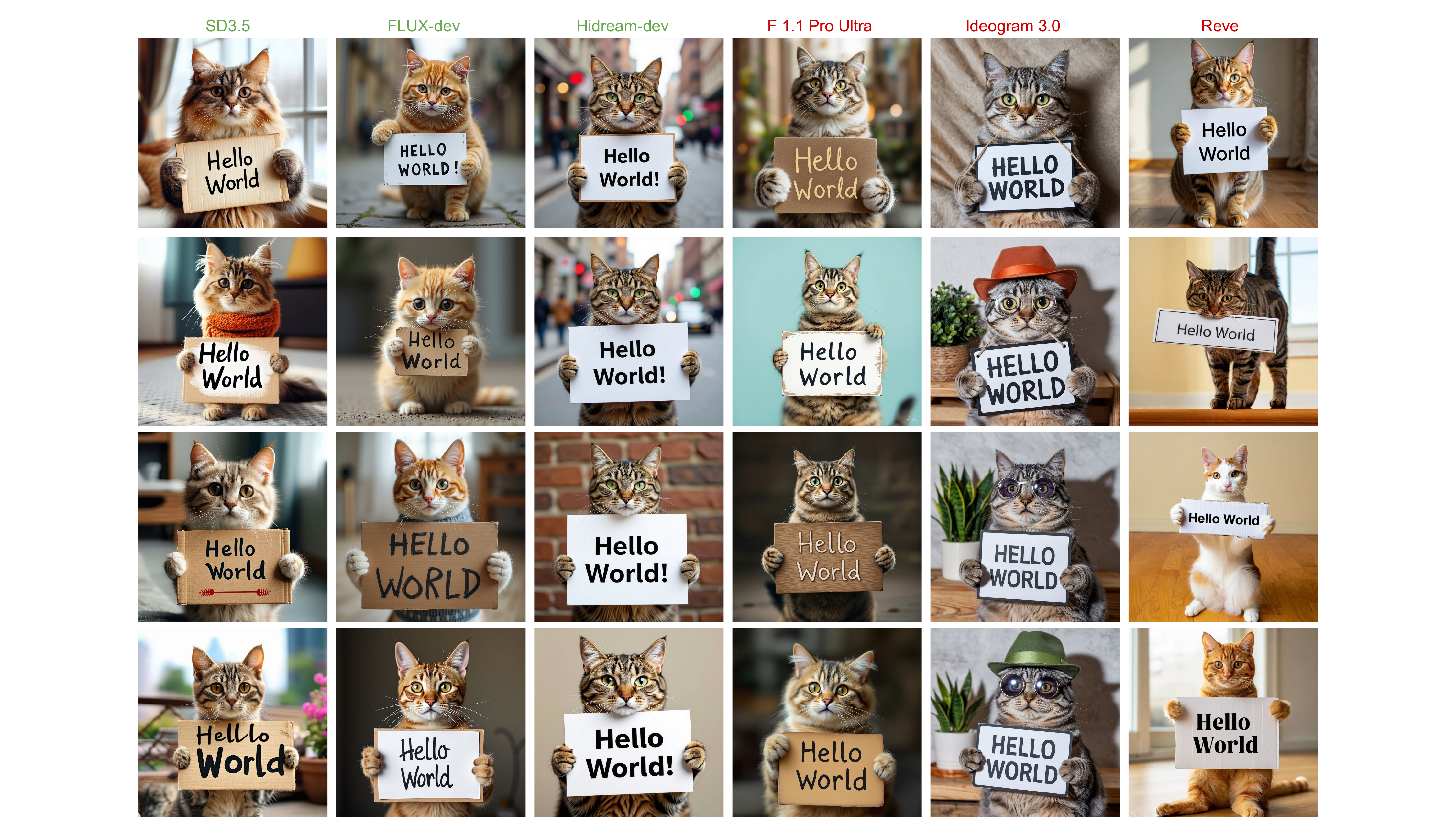}
    \includegraphics[trim=8cm 0cm 8cm 0cm, clip,width=1.0\textwidth]{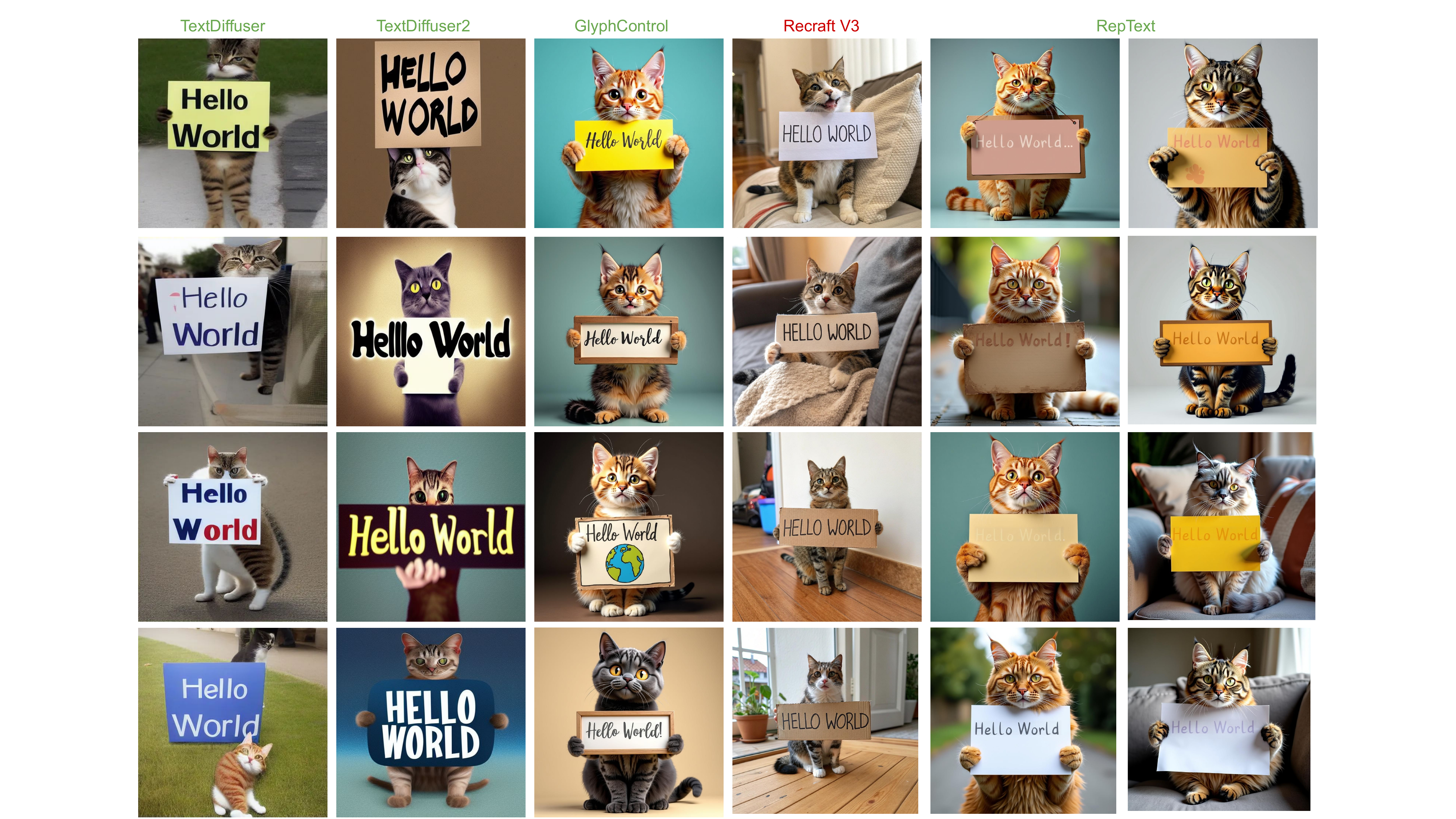}
    \caption{Comparison with \textcolor{green!50}{open-sourced} and \textcolor{red}{close-sourced} on monolingual rendering.}
    \label{fig:compare-monolingual}
\end{figure*}

\begin{figure*}[htbp]
    \centering
    \includegraphics[trim=8cm 0cm 8cm 1cm, clip,width=1.0\textwidth]{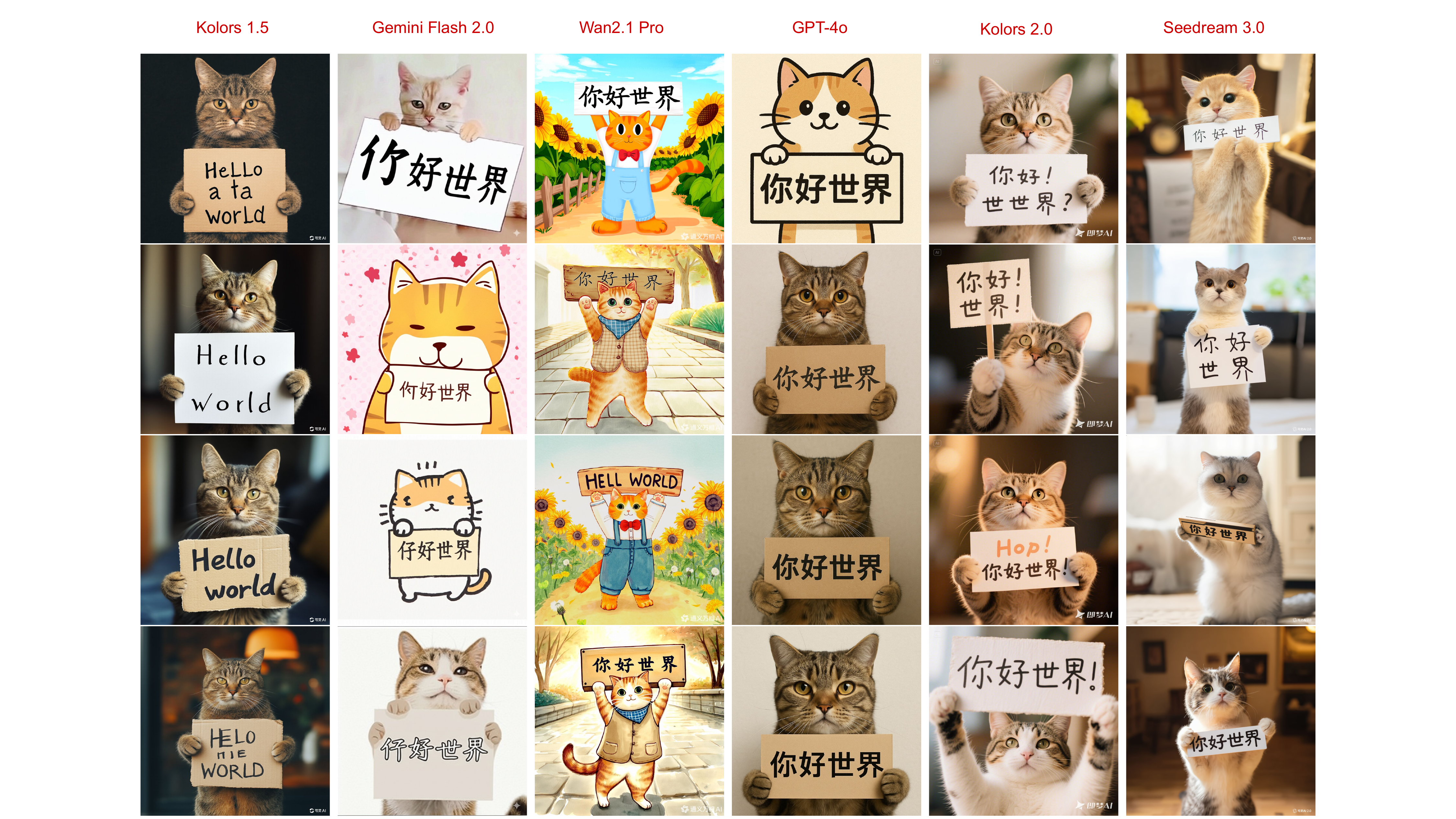}
    \includegraphics[trim=8cm 1cm 8cm 0cm, clip,width=1.0\textwidth]{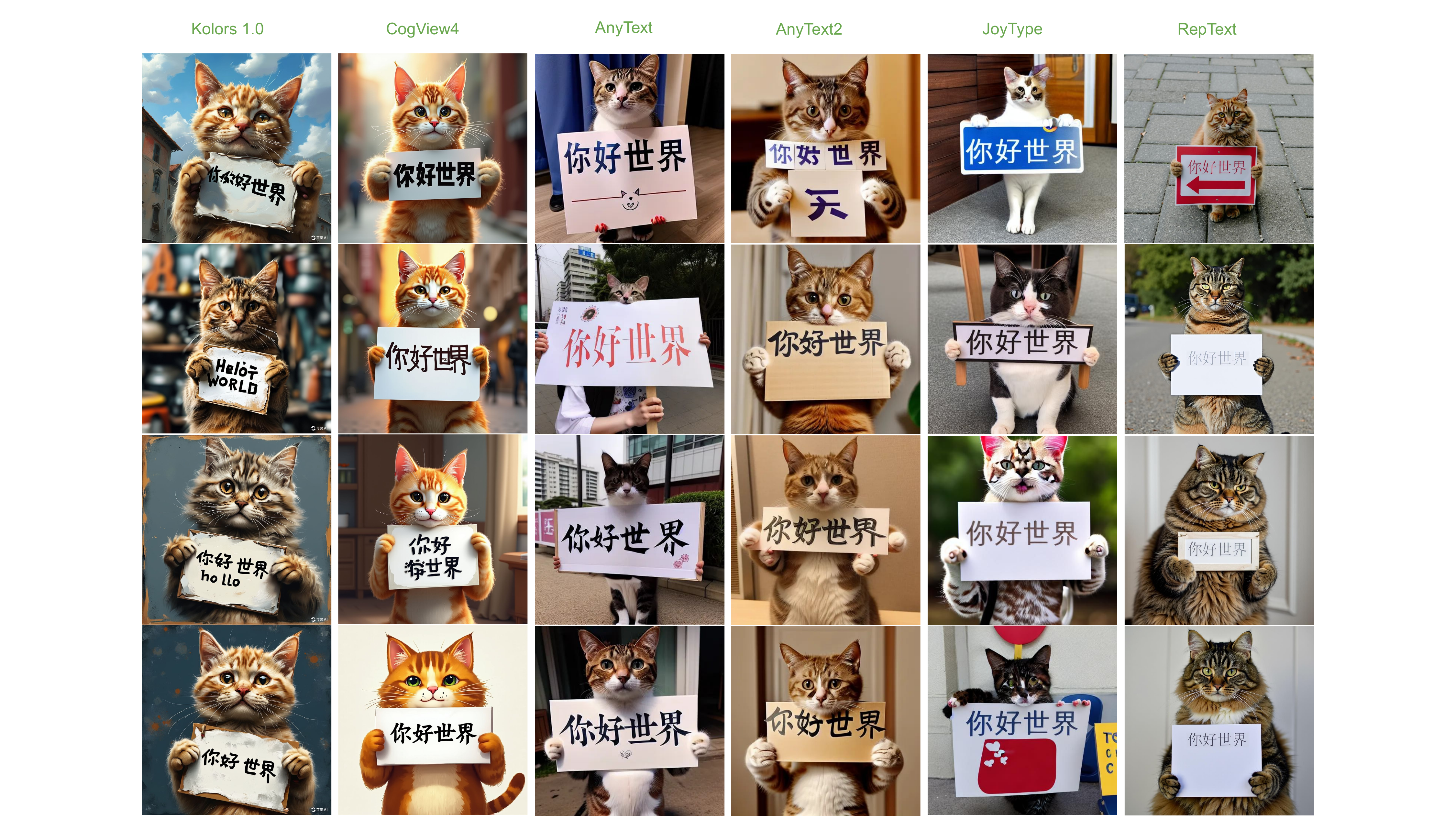}
    \caption{Comparison with \textcolor{green!50}{open-sourced} and \textcolor{red}{close-sourced} on multilingual rendering.}
    \label{fig:compare-multilingual}
\end{figure*}

For a comprehensive comparison, we compare with existing methods with
monolingual (Fig \ref{fig:compare-monolingual}) or multilingual (Fig \ref{fig:compare-multilingual}) text rendering ability.

\clearpage
\newpage

\subsection{Compatibility to Existing Works}

\begin{figure*}[h!]
    \centering
    \includegraphics[trim=6cm 2cm 6cm 2cm, clip,width=1.0\linewidth]{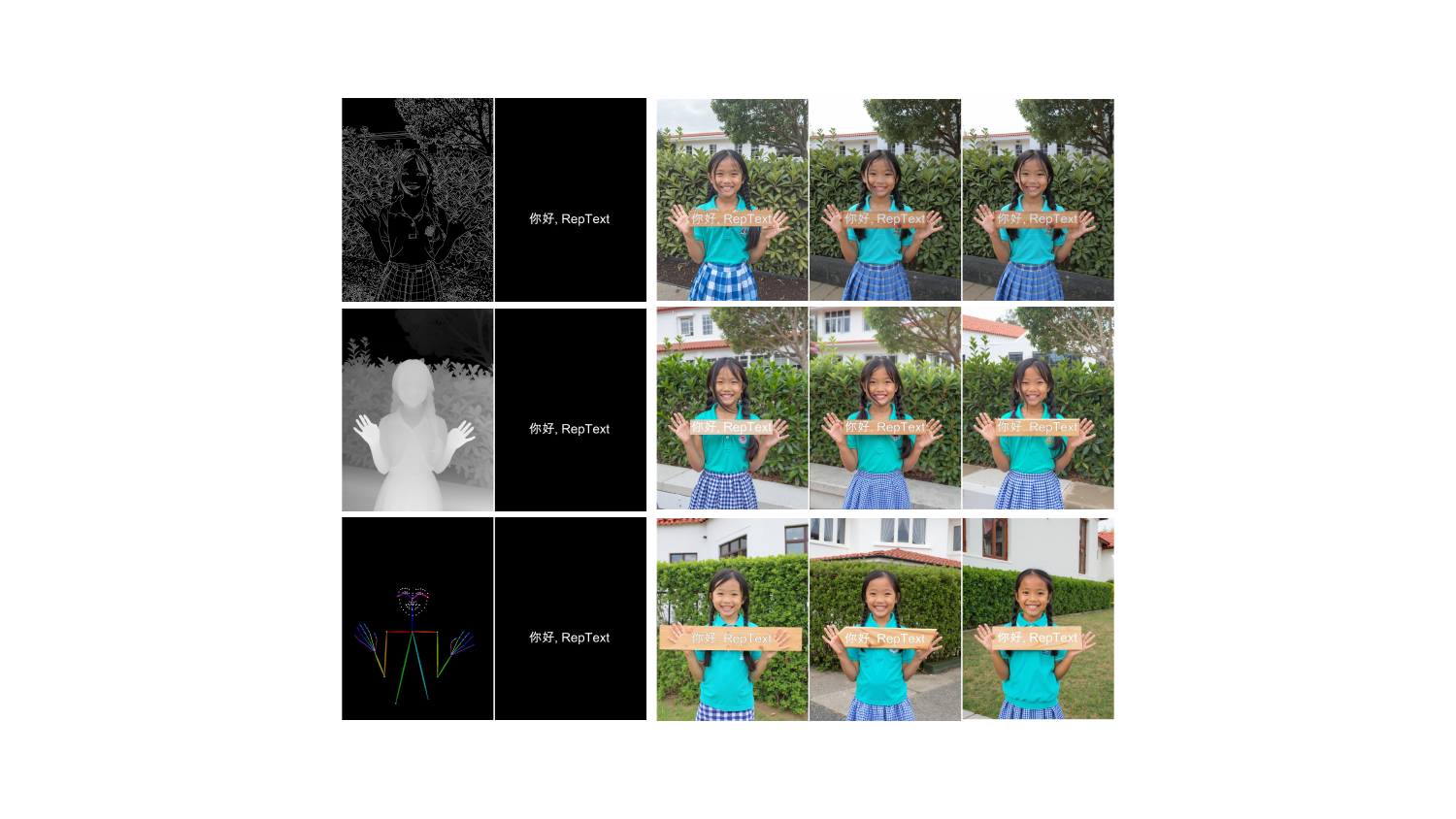}
    \includegraphics[trim=3cm 2cm 3cm 2cm, clip,width=1.0\linewidth]{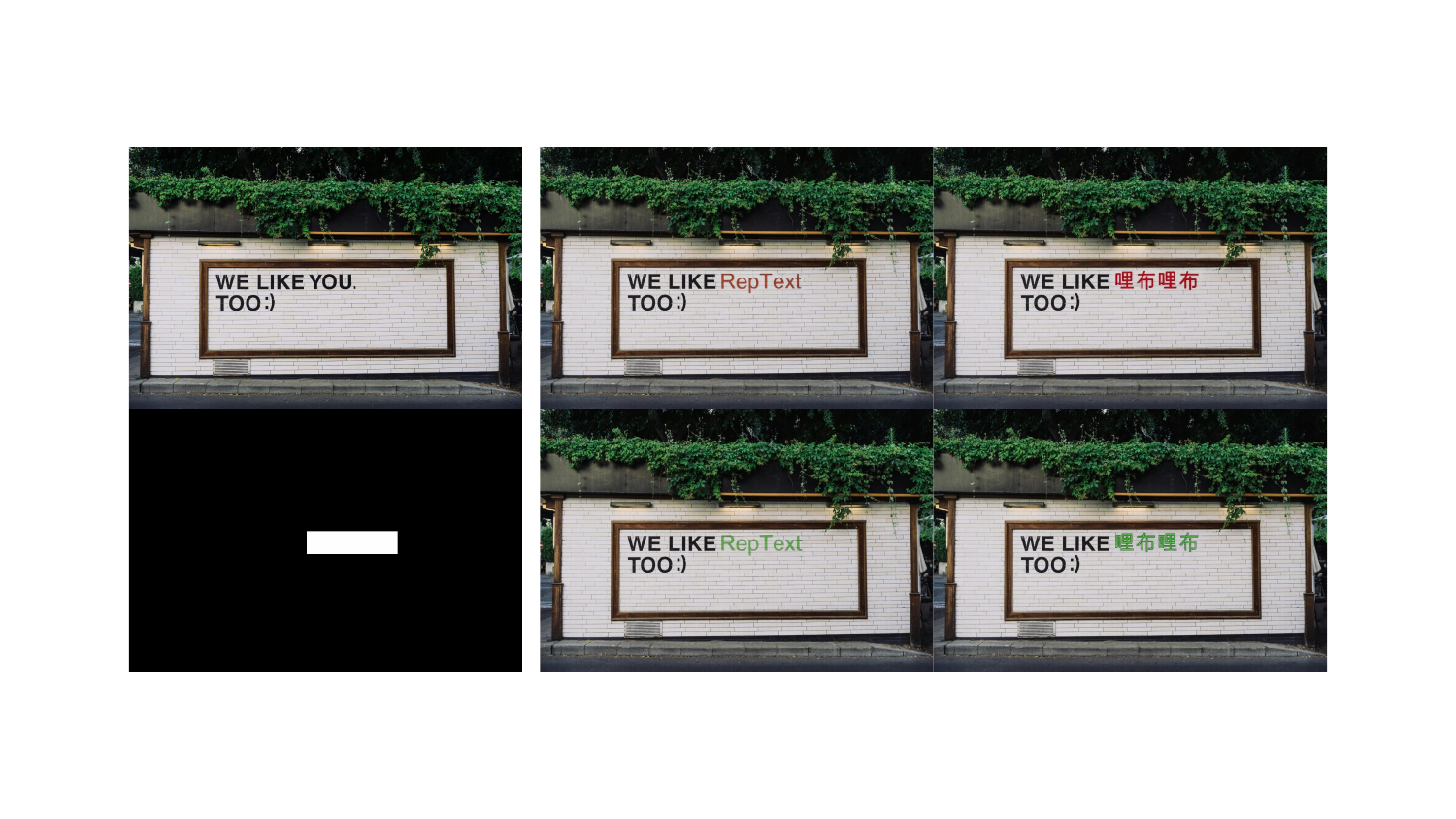}
    \caption{RepText with ControlNet-Union and ControlNet Inpainting.}
    \label{fig:controlnet}
\end{figure*}

\begin{figure*}[h!]
    \centering
    \includegraphics[trim=3cm 2cm 3cm 2cm, clip,width=1.0\linewidth]{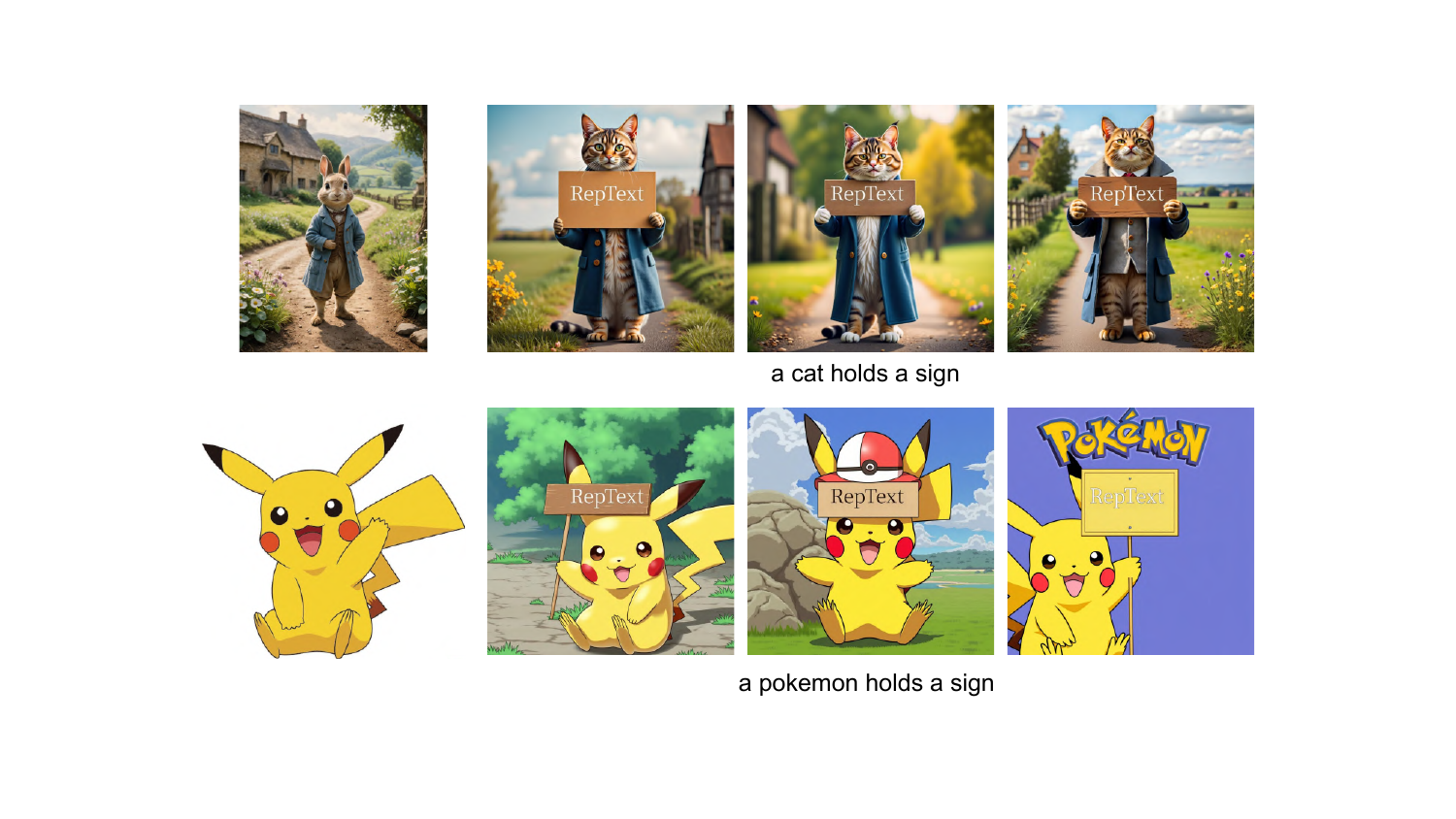}
    \caption{RepText with IP-Adapter.}
    \label{fig:ipa}
\end{figure*}

\clearpage
\newpage
\subsection{Ablation Studies}

\begin{figure*}[htbp]
    \centering
    \includegraphics[trim=0cm 1cm 1cm 0cm, clip,width=1.0\linewidth]{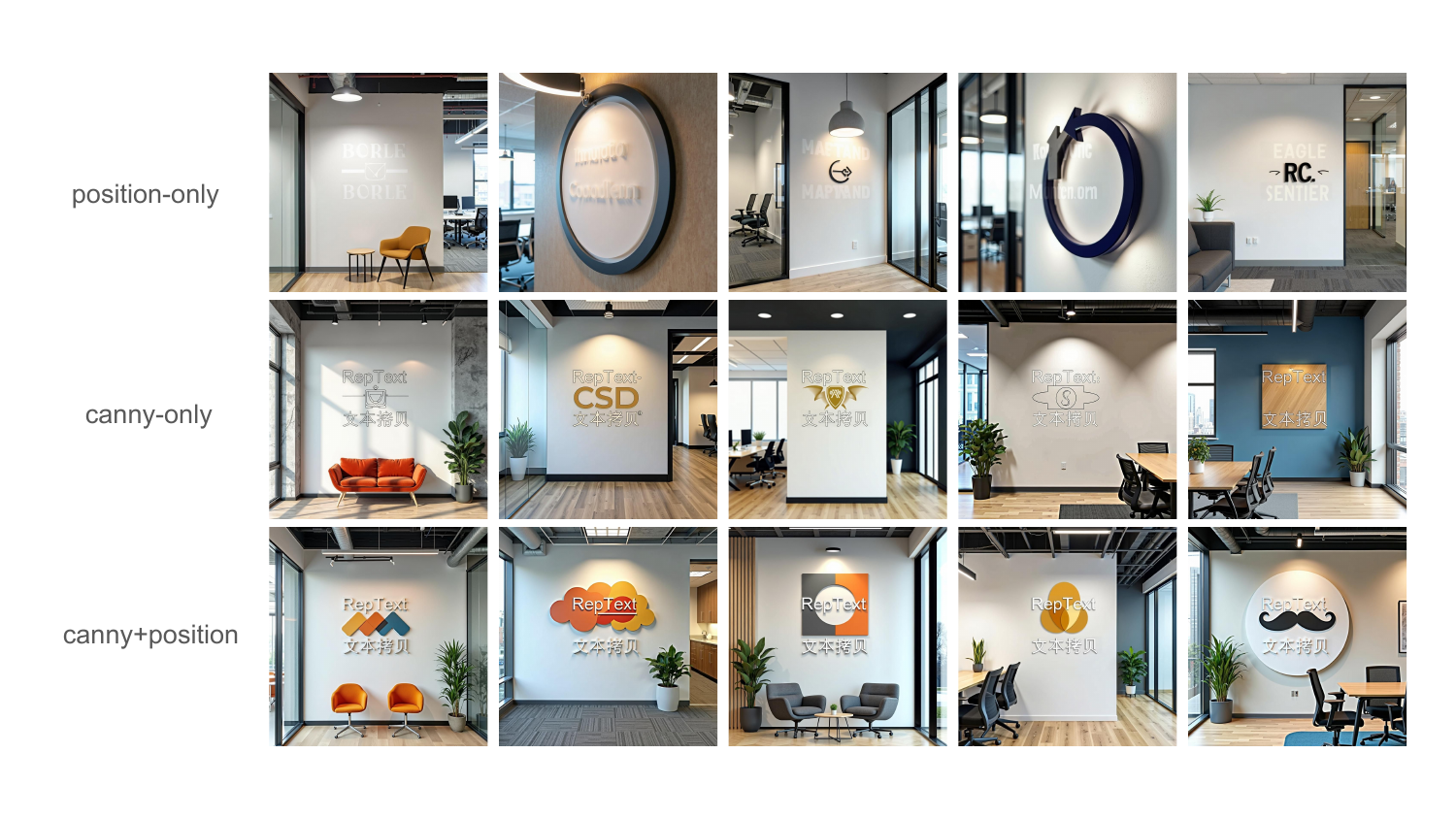}
    \caption{Ablation on the choice of control conditions.}
    \label{fig:ablation-choice-cn}
\end{figure*}

\begin{figure*}[htbp]
    \centering
    \includegraphics[trim=0cm 0cm 0cm 0cm, clip,width=1.0\linewidth]{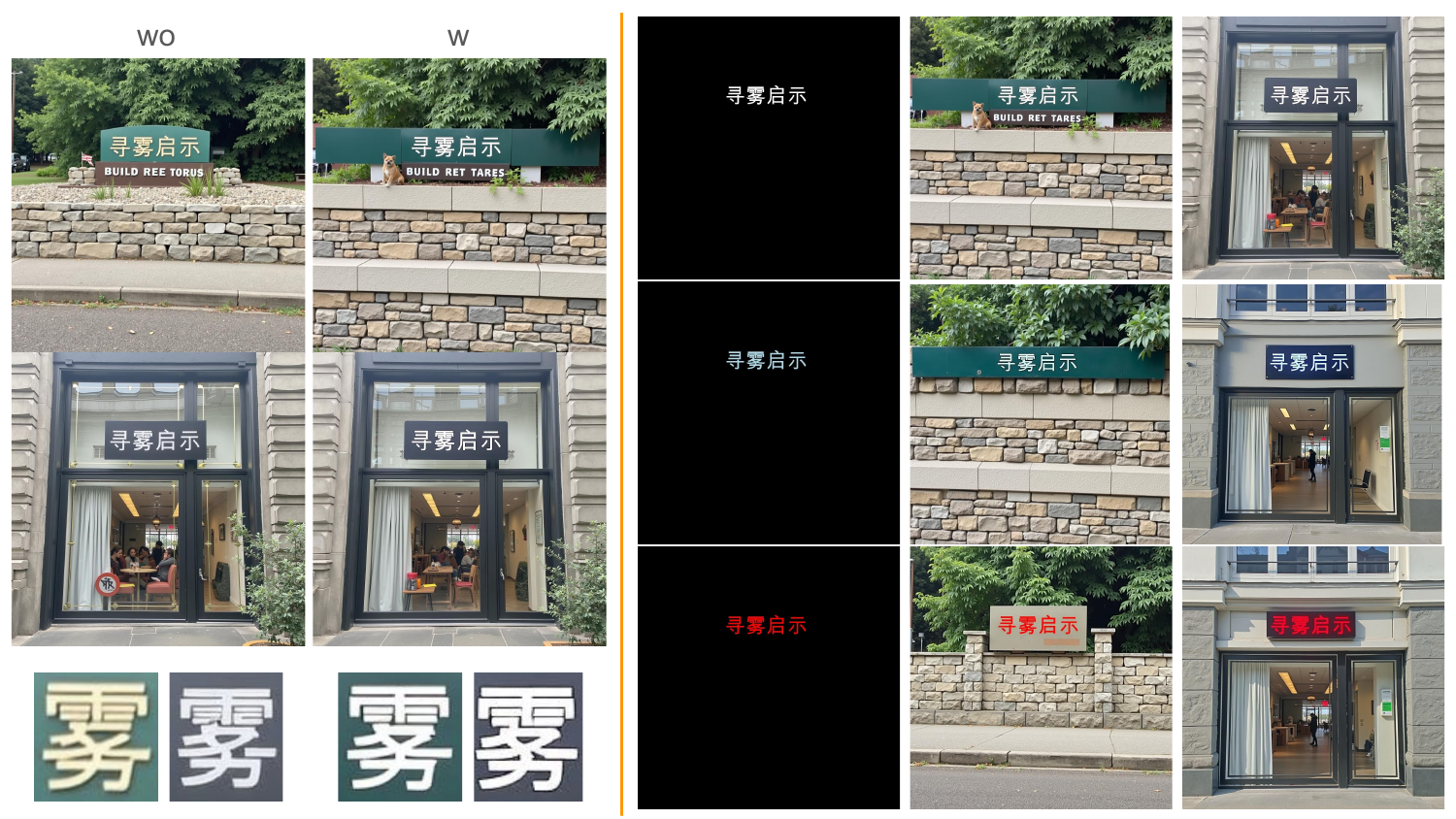}
    \caption{Ablation on the effect of glyph latent replication.}
    \label{fig:ablation-glyph-latent}
\end{figure*}

\begin{figure*}[htbp]
    \centering
    \includegraphics[trim=0cm 1cm 1cm 0cm, clip,width=1.0\linewidth]{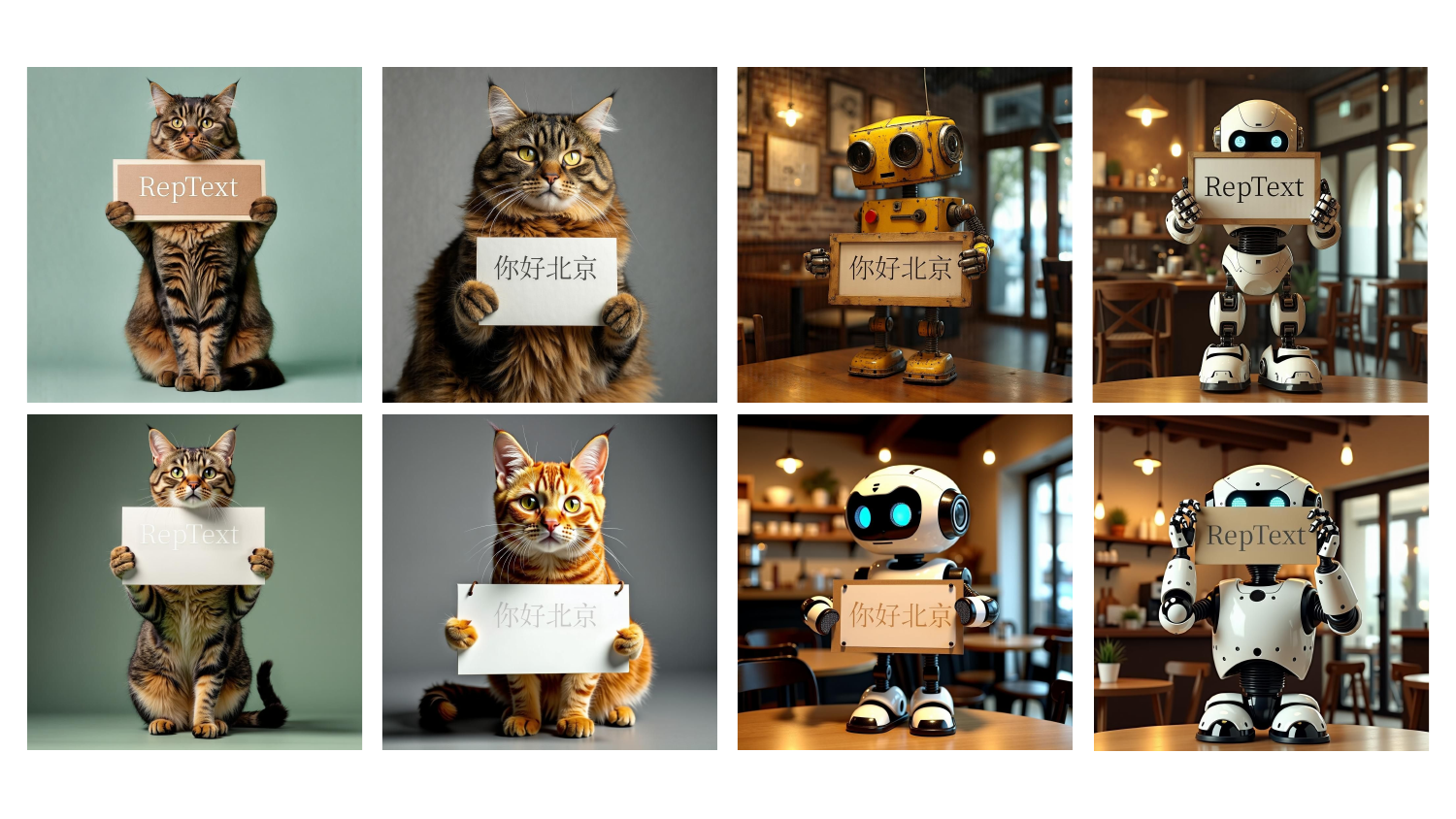}
    \caption{Ablation on the effect of regional masks. With regional mask (bottom), the generated images show better quality than direct injection (Top). Zoom in for better visualization.}
    \label{fig:ablation-region-mask}
\end{figure*}

\clearpage

\newpage
\subsection{Bad Cases}
\begin{figure*}[htbp]
    \centering
    \includegraphics[trim=5cm 2cm 5cm 2cm, clip,width=1.0\linewidth]{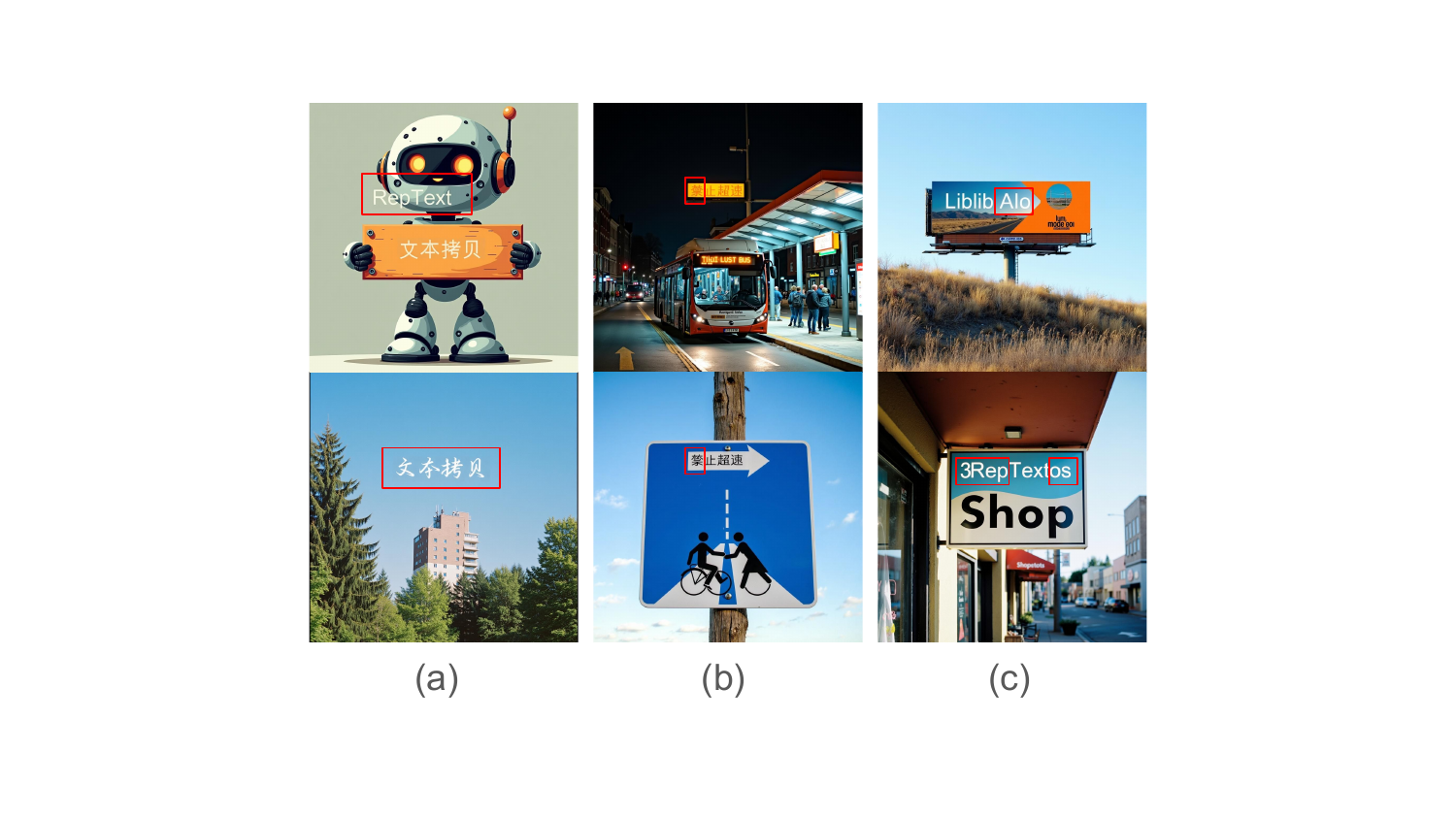}
    \caption{Typical Failure cases of RepText. (a) Disharmony with the scene, the text is rendered as signature or watermark. (b) Limited text accuracy especially for complex word or small font. (c) Extra texts appear as artifacts around text regions. The defective regions are highlighted by red box.}
    \label{fig:bad-cases}
\end{figure*}

We show several typical failure cases of RepText in Fig~\ref{fig:bad-cases}.

\end{document}